\documentclass[10pt,journal,compsoc]{IEEEtran}
\ifCLASSOPTIONcompsoc
  \usepackage[nocompress]{cite}
\else
  \usepackage{cite}
\fi
\ifCLASSINFOpdf
\else
\fi
\usepackage{epsfig}
\usepackage{graphicx}
\usepackage{amsmath}
\usepackage{amssymb}
\usepackage{bbm}
\usepackage{multirow}
\usepackage{color,soul}
\usepackage[dvipsnames]{xcolor}

\newcommand{\norm}[1]{\left\lVert#1\right\rVert}

\usepackage[pagebackref=true,breaklinks=true,letterpaper=true,colorlinks,bookmarks=false]{hyperref}

\definecolor{cgray}{rgb}{0.5, 0.5, 0.5}
\newcommand{\darkgrayed}[1]{\textcolor{cgray}{#1}}
\makeatletter
\newcommand*\titleheader[1]{\gdef\@titleheader{#1}}
\AtBeginDocument{
  \let\st@red@title\@title
  \def\@title{
    \vskip-1.3em
    \bgroup\normalfont\small\centering\@titleheader\par\egroup
    \vskip1.5em\st@red@title}
}
\makeatother

\titleheader{\darkgrayed{Accepted at IEEE Transactions on Pattern Analysis and Machine Intelligence, 2021.
\copyright IEEE}}

\title{HandVoxNet++: 3D Hand Shape and Pose Estimation using Voxel-Based Neural Networks} 

\begin{document}

\author{
$\;\;\;\;\;$Jameel Malik$^{1,2,3}$ $\;\;\;\;\;\;\;\;\,$  
Soshi Shimada$^{5,6}$ $\;\;\;\;\;\;\;\;\,$
Ahmed Elhayek$^{4}$ $\;\;\;\;\;\;\,$
\vspace{2pt}\\
$\;\;\;\;\,$Sk Aziz Ali$^{1,2}$  $\;\;\,$
$\;\;\, $Christian Theobalt$^{5,6}$ $\;\;\;\,$
Vladislav Golyanik$^{5,6}$ $\;\;\;\;$
Didier Stricker$^{1,2}$
\vspace{5pt}\\
$^{1}$TU Kaiserslautern$\;\;\;\;$
$^{2}$DFKI$\;\;\;\;$
$^{3}$NUST Pakistan$\;\;\;\;$
$^{4}$UPM Saudi Arabia$\;\;\;\;$
\vspace{2pt}\\
$^{5}$MPI for Informatics$\;\;\;\;$
$^{6}$Saarland Informatics Campus$\;\;\;\;$
}

\IEEEtitleabstractindextext{%
\begin{abstract} 
3D hand shape and pose estimation from a single depth map is a new and challenging computer vision problem with many applications. 
Existing methods addressing it directly regress hand meshes via 2D convolutional neural networks, which leads to artefacts due to perspective distortions in the images. 
To address the limitations of the existing methods, we develop HandVoxNet++, \textit{i.e.,} a voxel-based deep network with 3D and graph convolutions trained in a fully supervised manner. 
The input to our network is a 3D voxelized-depth-map-based on the truncated signed distance function (TSDF). 
HandVoxNet++ relies on two hand shape representations. 
The first one is the 3D voxelized grid of hand shape, which does not preserve the mesh topology and which is the most accurate representation. 
The second representation is the hand surface that preserves the mesh topology. 
We combine the advantages of both representations by aligning the hand surface to the voxelized hand shape either with a new neural \textit{Graph-Convolutions-based Mesh Registration} (GCN-MeshReg) or classical segment-wise \textit{Non-Rigid Gravitational Approach} (NRGA++) which does not rely on training data. 
In extensive evaluations on three public benchmarks, \textit{i.e.,}  SynHand5M, depth-based HANDS19 challenge and HO-3D, the proposed  HandVoxNet++ achieves state-of-the-art performance. 
In this journal extension of our previous approach presented at CVPR 2020, we gain $41.09\%$ and $13.7\%$ higher shape alignment accuracy on SynHand5M and HANDS19 datasets, respectively. 
Our method is \textbf{ranked first} on the HANDS19 challenge dataset (\textit{Task 1: Depth-Based 3D Hand Pose Estimation}) at the moment of the submission of our results to the portal in August 2020. 
\end{abstract} 

}

\maketitle

\IEEEdisplaynontitleabstractindextext
\IEEEpeerreviewmaketitle

\IEEEraisesectionheading{\section{Introduction}\label{sec:introduction}}
\label{sec:Intro}
\IEEEPARstart{T}{racking} and reconstruction of human hand pose in 3D is an extensively studied  computer vision problem which often arises in user authentication, augmented and virtual reality, gaming, movie production as well as human performance capture and analysis, among other fields \cite{supanvcivc2018depth,yuan2018depth,mueller2018ganerated,malik2020deepairsig}. 
Accurate 3D hand tracking can facilitate gesture recognition and enable new interfaces for human-computer interaction. 
While there are various accurate neural methods for 3D hand pose estimation focusing on RGB and depth  images \cite{yuan2018depth,xiong2019a2j,moon2017v2v,rad2018feature}, simultaneous estimation of 3D hand pose and 3D shape from a single depth map is an emerging research direction. 

\begin{figure*}[!ht]
\centering
\includegraphics[width=1\linewidth]{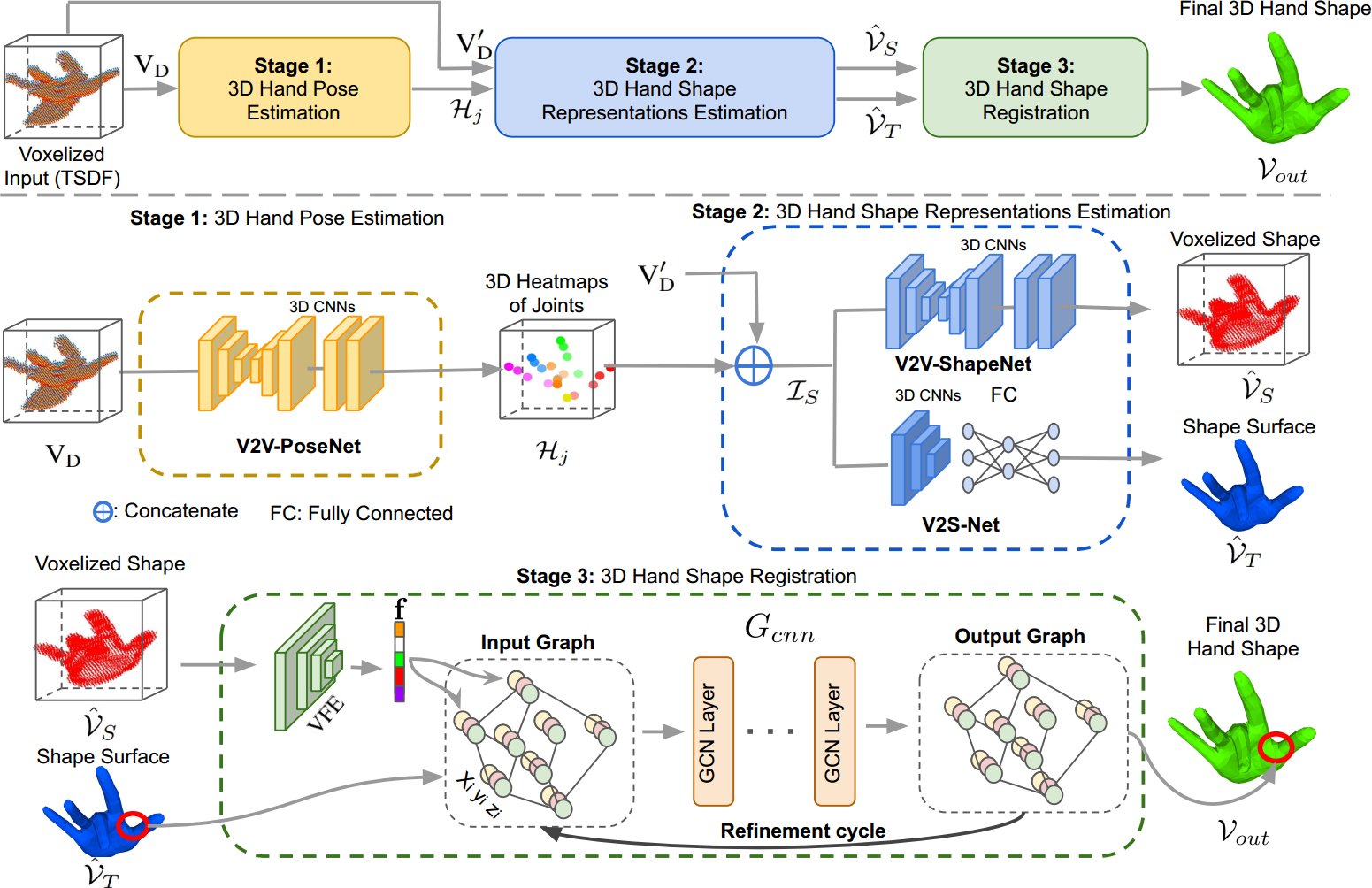}
\vspace{-6mm}
\caption{
\textbf{Overview of our approach for 3D hand shape and pose recovery.} 
The framework consists of three stages. 
In \textbf{Stage 1}, V2V-PoseNet accurately estimates 3D joints heatmaps $\mathcal{H}_j$ (\textit{i.e.,}~pose) from the TSDF-based 3D voxelized depth map $\textrm{V}_\textrm{D}$. The final hand shape $\mathcal{V}_{out}$ is estimated by the following stages.
In \textbf{Stage 2}, V2V-ShapeNet and V2S-Net estimate the voxelized shape $\hat{\mathcal{V}}_S$ and shape surface $\hat{\mathcal{V}}_T$ using 3D-convolution-based neural networks, respectively. 
Finally, in \textbf{Stage 3}, graph-convolution-based mesh registration  (\textit{i.e.,}~GCN-MeshReg) accurately fits $\hat{\mathcal{V}}_T$ to $\hat{\mathcal{V}}_S$. 
In the registration phase, a voxel feature extractor (VFE) first extracts a feature vector which is connected to each node ($\textrm{X}_\textrm{i}$,$\textrm{Y}_\textrm{i}$,$\textrm{Z}_\textrm{i}$) of the input graph. The output graph provides the deformed 3D mesh vertices.
In the refinement cycle, this estimate is further improved in an iterative manner. 
}
\label{fig:Pipeline}
\vspace{-2mm}
\end{figure*}

This problem is challenging because annotating real images for 3D shapes is cumbersome, due to varying hand shapes, self-occlusions (resulting in missing data in the depth maps), high number of degrees of freedom (DOF) and self-similarity of hand parts. 
On the other hand, the dense 3D hand mesh is a richer representation which is, in many cases, more useful than the bare 3D joints  \cite{taylor2016efficient,romero2017embodied,malik2019whsp,Qian2020}.
Benefiting from the recent advances in neural machine learning, several algorithms for  simultaneous hand pose and shape estimation have been introduced  \cite{zhang2019end,mueller_siggraph2019,ge20193d,malik2019whsp,malik2019simple,malik2018deephps}. 
Malik~\textit{et al.}~\cite{malik2018deephps} developed a 2D Convolutional Neural Network (CNN)-based approach that estimates shapes directly from 2D depth maps. 
The recovered shapes suffer from artefacts due to the limited representation capacity of their hand model \cite{ge20193d,malik2019whsp}. 
The same problem can occur even by embedding a realistic statistical hand model (\textit{i.e.,}~MANO \cite{romero2017embodied} or HTML \cite{Qian2020}) inside a deep neural network \cite{ge20193d,zhang2019end}. 
In contrast to these model-based approaches \cite{zhang2019end,malik2018deephps},  the direct regression-based approach proposed by Ge \textit{et al.}~\cite{ge20193d}  uses a monocular RGB image and achieves  more accurate results.
Another recent technique with direct regression from a single depth image is \cite{malik2019whsp}. 
All of the approaches mentioned above treat depth maps as 2D signals and process them with 2D CNNs, even though depth maps intrinsically provide 2.5D data. 
Training a 2D CNN to estimate 3D hand pose or shape given 2D representation of a depth map is highly non-linear and results in perspective distortions in the estimated outputs \cite{moon2017v2v}. 
V2V-PoseNet \cite{moon2017v2v} is the first work that uses 3D voxelized grid of a depth map to regress 3D joints heatmaps and, thus, avoids perspective distortions. 
Extending this work for shape estimation by directly regressing 3D heatmaps of mesh vertices is not feasible in practice, due to extremely high memory requirements.
In this article, we propose the first---to the best of our knowledge---architecture with 3D CNNs and graph convolutions called HandVoxNet++, which simultaneously estimates 3D shape and 3D pose given a voxelized depth map, see Fig.~\ref{fig:Pipeline} for an overview. 
HandVoxNet++ is based on 3D and graph convolutions which regresses two different representations of hand shape, namely voxelized hand shape and hand surface   (Secs.~\ref{sec:method_overview}--\ref{sec:NetTraining}). 
The voxelized hand shape is estimated from a new  \textit{voxel-to-voxel} network relying on Truncated Signed Distance  Field (TSDF), and establishes a one-to-one mapping between the  voxelized depth map and the voxelized hand shape. 
Since the estimated voxelized shape does not preserve the hand mesh topology and the number of vertices, we also estimate hand surface with our \textit{voxel-to-surface} network. 
The voxel-to-surface network does not establish a one-to-one mapping  between the voxelized depth measurements and shapes; the accuracy of  the estimated hand surface is low but the hand topology is preserved. 
To combine the advantages of both representations, we register the  estimated hand surface to the estimated voxelized hand shape. 
In~\cite{HandVoxNet2020}, we introduced two different variants of shape registration  methods, \textit{i.e.,} CNN-based \cite{ShimadaDispVoxNets2019} and  NRGA-based \cite{Ali_NRGA_2018}. 
In this article, we propose two additional novel and more accurate  registration algorithms, namely GCN-MeshReg and NRGA++. 
In the GCN-MeshReg, we employ graph convolutions to explicitly utilize the topology of the hand shape. 
The \textit{Non-Rigid Gravitational Approach} (NRGA) for point set alignment \cite{Ali_NRGA_2018} is extended to support segmented hand surface. 
We call this extended version of the registration method NRGA++. 
Next, to increase the robustness and accuracy of the hand pose estimation, we perform 3D data augmentation on the voxelized depth maps (Sec.~\ref{ssec:DataAugmentation}). 
Our main contribution is the new state-of-the-art HandVoxNet++ approach for 3D hand shape and pose estimation from a single depth map. 
It includes a \textit{voxel-to-voxel} and \textit{voxel-to-surface} networks  with 3D CNNs as well as two adapted and improved mesh registration components. 
This work is an extension of our conference paper \cite{HandVoxNet2020}. 
Compared to \cite{HandVoxNet2020}, the new contributions in this article are: 
\begin{itemize} 
  \item A new TSDF-based \textit{voxel-to-voxel} network for 3D hand pose estimation (Sec.~\ref{ssec:PoseEstimation}). 
  Our TSDF-based representation of  depth map achieves $19.8$\% 
  improvement in accuracy compared to binary voxelized grid  representation used in \cite{moon2017v2v} (Sec.~\ref{ssec:EvalHandPose}). 
  \item The first work that applies graph convolutions to a registration problem between a hand mesh and a 3D voxelized hand shape (Sec.~\ref{sssec:DeepReg}). 
  Furthermore,  we  propose  a novel iterative refinement strategy for the shape registration. This allows to  significantly  improve the  accuracy and computational time compared to
  \cite{HandVoxNet2020} (Sec.~\ref{ssec:EvalShapeEstim}). 
  \item A new segment-wise point set alignment method NRGA++ which is ${\sim}150$ times faster than the NRGA proposed in \cite{HandVoxNet2020} (Sec.~\ref{sssec:GenReg} and Sec.~\ref{ssec:EvalShapeEstim}). 
  \item Extensive experiments on three recent benchmarks (\textit{i.e.,} SynHand5M, depth-based HANDS19 challenge, and HO-3D \cite{hampali2020honnotate}) illustrate that HandVoxNet++ achieves: 
   \begin{enumerate}
   \item Superior or comparable performance than the existing approaches, 
   \item More accurate hand shapes compared to the state-of-the-art HandVoxNet approach \cite{HandVoxNet2020}(Sec.~\ref{sec:experiments}) and 
   \item The first place on the task of ``depth-based 3D hand pose estimation of the HANDS19" challenge, at the moment of our submission to its web portal in August 2020. 
   \end{enumerate}
\end{itemize}

\section{Related Work}\label{sec:related_work} 

In this section, we review the most relevant existing methods for neural hand pose and shape estimation from RGB and depth images. 
\noindent\textbf{Neural Hand Pose and Shape Estimation.} 
Most approaches for 3D hand reconstruction and tracking from monocular RGB and depth images and event streams  estimate hand poses only, \textit{i.e.,} a sparse set of 3D hand joints 
\cite{cai20203d,park20203d,supanvcivc2018depth,yuan2018depth,mueller2018ganerated,Nehvi2021,rudnev2021eventhands}. 
The approaches which estimate hand shape and pose simultaneously are still in the minority. 
Malik \textit{et al.}~\cite{malik2018deephps} proposed the first deep neural network for hand pose and shape estimation from a single depth image. 
To this end, they developed a model-based hand pose and shape layer which is embedded inside their deep network. 
Their approach suffers from artefacts due to the difficulty in optimizing complex hand shape parameters inside the network. 
Ge \textit{et al.}~\cite{ge20193d} developed a direct regression-based algorithm for hand pose and shape estimation from a single RGB image. 
They highlight that the representation capacity of the statistical deformable hand model (\textit{i.e.,}~MANO~\cite{romero2017embodied}) could be limited due to the small amount of training data and the linear bases utilized for the shape recovery. 
\cite{zhang2019end,hasson2020leveraging,yang2020seqhand} introduced similar MANO hand-model-based neural approaches using monocular RGB image. 
Qian \textit{et al.}~\cite{Qian2020} came up with the first parametric hand texture model,  which can be applied in 3D hand reconstruction and personalization from a single image as well as a differential neural hand appearance layer. 
Next, a weakly-supervised neural approach using a single  depth image was presented in \cite{malik2019whsp}. 
\cite{baek2020weakly} directly regressed the hand mesh  vertices and a dense correspondence map using 2D fully  convolutional neural architecture. 
In \cite{Zhou_2020_CVPR}, many sources of  hand training data are utilized to train a deep learning architecture which estimates 3D pose given monocular RGB image. 
Thereafter, the 3D pose is used to estimate the shape parameters. 
All of the above-mentioned methods use 2D CNNs and treat the depth maps (the methods which support those) as 2D data. 
Consequently, the deep network is likely to produce perspective distortions in the shape and pose estimations \cite{moon2017v2v}. 
In contrast, we propose the first 3D-convolution-based architecture which effectively establishes a one-to-one mapping between the voxelized depth map and the voxelized hand shape. 
This one-to-one mapping allows to more accurately reconstruct the hand shapes. 

\noindent\textbf{Shape Registration.} 
Recently, Graph Convolutional Networks (GCNs) have received much attention in computer vision. The concept of GCNs was introduced in \cite{kipf2016semi} for semi-supervised classification. Later, it was  used to develop powerful solutions of many computer vision problems. For instance, Kolotouros \textit{et al.}~\cite{kolotouros2019cmr} apply graph-convolution-based architecture to regress human 3D shapes and poses given 2D RGB images and human template mesh. 
Tretschk \textit{et al.}~\cite{Tretschk2020DEMEA} demonstrate the effectiveness of graph convolutions for mesh autoencoding including hands from the SynHand5M dataset \cite{malik2018deephps}. 
A GCNs-based model for both hand and object pose estimation was proposed in \cite{hopenet}. 
Unlike other works, we propose an accurate and fast GCNs-based approach for the mesh to the voxelized shape registration with an iterative refinement strategy. 
On the other hand, fully-convolutional networks were shown to perform well in geometry regression tasks  \cite{shimada2019ismo,golyanik2018hdm,moon2017v2v,wu2016learning}. 
\noindent\textbf{Neural Hand Pose Estimation from Depth.} 
In general, deep learning-based hand pose estimation methods can be classified into two categories. 
The first one encompasses the discriminative methods which directly estimate hand joint locations using CNNs  \cite{chen2019so,cai2019exploiting,xiong2019a2j,poier2019murauer,rad2018feature,moon2017v2v,guo2017region,ge2017robust,malik20183dairsig}.  
The second category is hybrid methods which explicitly incorporate hand structure inside deep networks \cite{malik2018structure,wan2018dense,ge2018point,oberweger2017deepprior++,malik2017simultaneous}. 
The disriminative methods achieve higher accuracy compared to the hybrid methods. 
The \textit{voxel-to-voxel} approach \cite{moon2017v2v} is powerful and highly effective because it uses 3D convolutions to learn a one-to-one mapping between the 3D voxelized depth map and 3D heatmaps of hand joints.  
Notably, the voxelized representation of depth maps is best suited for 3D data augmentation to improve the robustness and accuracy of the estimations.
A few methods perform data augmentation on depth maps~\cite{oberweger2017deepprior++,tompson2014real} or voxelized depth maps~\cite{moon2017v2v}.
In this work, we integrate the \textit{voxel-to-voxel} approach with our pipeline and, additionally, perform new 3D data augmentation on voxelized depth maps that further improves the 3D pose estimation.      
The \textit{voxel-to-voxel} approach transformed 2D depth map to 3D binary voxelized grid representation which looses information in the 3D point cloud due to the binary quantization. 
Ge \textit{et al.}~\cite{ge2019handshapepose} utilized the TSDF representation of depth map however, they directly regressed the hand joint positions using a neural architecture which does not establish a one-to-one relation between the depth map and the pose. 
In contrast, we propose to establish a one-to-one mapping between the TSDF-based 3D voxelized grid of 2D depth map and 3D heatmaps of joint positions. 
The TSDF representation better encapsulates the information of the 3D point cloud and, thereby, significantly enhances the accuracy of 3D hand pose estimation. 

\section{Method Overview}
\label{sec:method_overview}
Given an input depth map, our goal is to estimate \textrm{N} 3D hand joint locations $\mathcal{J} \in \mathcal{R}^{3 \times \textrm{N}}$ (\textit{i.e.,}~3D pose) and $\textrm{K}$ 3D vertex locations $\mathcal{V} \in \mathcal{R}^{3 \times \textrm{K}}$ (\textit{i.e.,}~3D shape). 
Fig.~\ref{fig:Pipeline} shows an overview of the proposed approach. 
The input is transformed into a voxelized grid (\textit{i.e.,}~$\textrm{V}_{\textrm{D}}$) of size $88\times88\times88$, by using intrinsic camera parameters, a fixed cube size and the truncated signed distance function (Section~\ref{ssec:PoseEstimation}). 
For hand pose estimation (\textit{i.e.,} Stage 1), $\textrm{V}_\textrm{D}$ is provided as an input to the \textit{voxel-to-voxel} pose regression network (\textit{i.e.,}~V2V-PoseNet) that directly estimates 3D joint heatmaps $\{\mathcal{H}_j\}_{j=1}^\textrm{N}$. 
Each 3D joint heatmap is represented as $44\times44\times44$ voxelized grid.
The input of the shape estimation network  (\textit{i.e.,} stage 2 ) is the concatenation of $\mathcal{H}_j$ (\textit{i.e.,} the output of Stage 1) and $\textrm{V}_\textrm{D}$. We call this concatenated input as $\mathcal{I}_S$. To this end, we resize $\textrm{V}_\textrm{D}$ to $44\times44\times44$ voxel grid size (\textit{i.e.,} ${\textrm{V}}'_\textrm{D}$). 
The voxelized hand shape (\textit{i.e.,}~$64\times64\times64$ binary grid) is directly regressed via 3D CNN-based \textit{voxel-to-voxel} shape regression network (\textit{i.e.,}~V2V-ShapeNet), by using $\mathcal{I}_S$ as an input. Notably, V2V-ShapeNet establishes a one-to-one mapping between the voxelized depth map and the voxelized shape. Therefore, it produces an accurate voxelized shape representation but does not preserve the topology of hand mesh and the number of mesh vertices. 
To regress hand surface, $\mathcal{I}_S$ is fed to the 3D CNN-based \textit{voxel-to-surface} regression network (\textit{i.e.,}~V2S-Net). Since the mapping between
$\mathcal{I}_S$ and hand surface is not one-to-one, it is therefore less accurate. 
To combine the advantages of the two hand shape representations, Stage 3 registers the estimated hand surface to the estimated voxelized hand shape. 
The voxelized shape is provided as an input to a 3D-convolution-based network that produces a feature vector \textrm{f} of size = $2187$. 
\textrm{f} is concatenated to each node of the hand mesh. This representation is fed to the graph convolution neural networks as input, and the networks return the registered hand shape that matches the voxelized hand shape. We repeat this procedure using the output from the graph convolution networks as a new template hand shape, which we term {refinement cycle}. Thanks to this refinement cycle, we obtain highly accurate and smooth hand shape as the final output.
Please note that Fig.~\ref{fig:Pipeline} is based on GCN-MeshReg as it enables the best performance. 
However, Stage 3 can be replaced with other registration approaches such as DispVoxNet, NRGA, or NRGA++. 

\section{The Proposed HandVoxNet Approach}
\label{sec:HandVoxNet}
In this section, we explain our proposed HandVoxNet++ approach by highlighting the function and effectiveness of each of its components. 
We develop an effective solution that produces reasonable hand shapes via voxel-based convolutional networks. 
To this end, our approach exploits the estimated 3D heatmaps of hand joints as a strong pose prior and voxelized depth map to accurately estimate the hand shape representations. 
Moreover, our 3D data augmentation on voxelized depth maps 
allows to further improve the accuracy and robustness of 3D hand pose
estimation. 
\subsection{{Volumetric Input Representation}} 
\label{ssec:PoseEstimation} 
In this section, we discuss two possible conversions of the depth map represented in 2D image to 3D volume; namely occupancy grid and the TSDF-based grid. 
The main motivation for this modification is that the depth map representation of 3D hand distort the hand during the projection from 3D space to the 2D image space. 
This makes the training process more challenging as the network receives as input a distorted representation of the real hand. 
Another drawback of the depth map representation is the highly non-linear mapping between it and the 3D output which complicates the learning process~\cite{moon2017v2v}. 

To overcome these limitations, 
the input depth map should be encoded into a volumetric representation. The easiest  volumetric representation is the occupancy (binary) grid~\cite{moon2017v2v}. 
Although this representation  allows to overcome the depth maps limitations, {it  cannot differentiate voxels behind and in front of the observed surface}. The reason of this fact is that it is generated based on incomplete information about the 3D hand shape (\textit{i.e.,} the depth map which only captures the observed surface from the view of camera). 

We encode the depth map into 3D voxelized grid representation using the TSDF. 
The input depth map is first converted into a set of 3D points, and these points are further discretized in the range $[$$1$, $88$$]$ (Sec.~\ref{sec:NetTraining}).    
Then, a 3D voxelized grid of size $\textrm{88} \times \textrm{88} \times \textrm{88}$ (\textit{i.e.,} $\textrm{V}_\textrm{D}$) can be created by calculating \textrm{V}(\textit{k}) for each voxel \textit{k}. In the TSDF representation, \textrm{V}(\textit{k}) can be obtained as: 
\begin{equation} \label{eq:tsdf}
\textrm{V}(k) = \textrm{min}(\textrm{max}(\textrm{d}(\textit{k}_c)/\mu , -1), +1), 
\end{equation}
where $\textrm{d}(\textit{k}_c)$ is the signed distance from the voxel center $\textit{k}_c$ to the nearest surface point in the set of discretized 3D points. We use the standard Euclidean distance. The sign of  $\textrm{d}(\textit{k}_c)$ is positive if the depth values of $\textit{k}_c$ is less than the depth value of the nearest surface point otherwise, the sign is negative. We use $\mu$ = $3$ as the truncated distance value. In accurate TSDF representation, the nearest surface point is found by checking all the 3D points which is highly time consuming and therefore, not suitable for real-time applications. The projective TSDF \cite{newcombe2011kinectfusion} is practically more feasible because the nearest surface point is calculated on the line of sight in the camera frame. However, the projective TSDF is an approximation of the accurate TSDF and consequently, it loses some information \cite{ge2018real}.
In the projective directional TSDF reprensentation \cite{song2016deep}, 
the voxel value stores the 3D offset vector (\textit{i.e.,} $[dx,dy,dz]$) to the nearest point. 
In the experiment section \ref{sec:experiments}, we demonstrate that the projective TSDF representation yields the best performance in our case and effectively establishes a one-to-one mapping between the TSDF-based voxelized depth map and the 3D heatmaps of the joints.

\subsection{3D Hand Shape Estimation} 
\label{ssec:ShapeEstimation} 
As aforementioned, estimating 3D hand shape from a 2D depth map by using 2D CNN is a highly non-linear mapping. 
It compels the network to perform perspective-distortion-invariant estimation which causes difficulty in learning the shapes. To address this challenge, we develop a fully voxel-based deep network that effectively utilizes the estimated 3D pose and voxelized depth map to produce reasonable 3D hand shapes. Our proposed approach for 3D shape estimation comprises of two main phases. In the first phase, we estimate the shape surface and the voxelized hand shape. In the second phase, we register the estimated shape surface to the estimated voxelized hand shape. We discuss several registration approaches and provide their comparative analysis.  

\noindent\textbf{Voxelized Shape Estimation.}
Our idea is to estimate 3D hand shape in the voxelized form via 3D CNN-based network. It allows the network to estimate the shape in such a way that minimizes the chances for perspective distortion. Inspired by the approach proposed in the recent work \cite{malik2019whsp}, we consider sparse 3D joints as the latent representation of dense 3D shape. However, in this work, we combine 3D pose with the depth map which helps to represent the shape of hand more accurately. 
Furthermore, here we use more accurate and useful representations of 3D pose and 2D depth image which are 3D joints heatmaps and a voxelized depth map, respectively. The V2V-ShapeNet module is shown in Fig.~\ref{fig:Pipeline}. {It can be considered as the 3D shape decoder:} \begin{equation} \label{eq:0}
\hat{\mathcal{V}}_S \sim \textit{Dec}( \mathcal{H}_j \oplus {\textrm{V}}'_\textrm{D}) = p(\mathcal{V}_S |\mathcal{I}_S),
\end{equation} 
where  $p(\mathcal{V}_S |\mathcal{I}_S)$ is the decoded distribution. {The decoder $\textit{Dec}(\cdot)$} learns to reconstruct the voxelized hand shape $\hat{\mathcal{V}}_S$ as close as possible to the ground truth voxelized hand shape $\mathcal{V}_S$.
The V2V-ShapeNet is a 3D CNN-based architecture~\cite{HandVoxNet2020} that directly estimates the probability of each voxel in the voxelized shape indicating whether it is the background (\textit{i.e.,}~$0$) or the shape voxel (\textit{i.e.,}~$1$). 
The per-voxel binary cross entropy loss $\mathcal{L}_{\mathcal{V}_S}$ for voxelized shape reconstruction reads: 
\begin{equation} \label{eq:1}
\mathcal{L}_{\mathcal{V}_S} = - (\mathcal{V}_S\hspace{1mm}  \log(\hat{\mathcal{V}}_S) + (1-\mathcal{V}_S) \hspace{1mm}\log(1-\hat{\mathcal{V}}_S)),
\end{equation}
where $\mathcal{V}_S$ and $\hat{\mathcal{V}}_S$ are
the ground truth and the estimated voxelized hand shapes, respectively. 

\noindent\textbf{Shape Surface Estimation.} 
The hand poses of the shape surfaces and voxelized shapes need to be similar for an improved shape registration. 
To facilitate the registration, 
we employ V2S-Net deep network~\cite{HandVoxNet2020} which directly regresses $\mathcal{V}$. Based on the similar concept of hand shape decoding (as mentioned before), $\mathcal{I}_S$ is provided as an input to this network while the decoded output is the reconstructed hand mesh (see Fig.~\ref{fig:Pipeline}). The hand shape surface reconstruction loss $\mathcal{L}_{\mathcal{V}_T}$ is given by the standard Euclidean loss as: 
\begin{equation} \label{eq:3}
\mathcal{L}_{\mathcal{V}_T} = \frac{1}{2}\norm{ \hat{\mathcal{V}}_T - \mathcal{V}_T }^2, 
\end{equation}
where $\mathcal{V}_T$ and $\hat{\mathcal{V}}_T$ are the respective ground truth and reconstructed hand shape surfaces. 

\subsection{Shape Registration} 
\label{ssec:ShapeRegistraion}
As mentioned above, V2S-Net can estimate hand shapes while preserving the order and number of points. 
Unfortunately, as the V2S-Net uses fully connected (FC) layer for mesh vertices regression, it looses local spatial information. 
Although, estimating the voxelized hand shape by 3D convolutional layers  guarantees a one-to-one mapping between the input and the output, it results in an inconsistent number of points and loses point order. 
To preserve the hand shape topology while relying on voxelized hand shape representation, 
we register the shape estimated by V2S-Net to the probabilistic  shape representation estimated by V2V-ShapeNet. 
Therefore, we propose a new neural shape registration approach {GCN-MeshReg} (Sec.~\ref{sssec:DeepReg}) along with a classical optimization-based segmentwise non-rigid gravitational approach which we abbreviate as NRGA++ (Sec.~\ref{sssec:GenReg}). 
Our neural shape registration algorithm not only achieves higher accuracy compared to the classical optimization-based method but is also much faster. 
On the downside, compared to NRGA++, it requires a training dataset for hand shape alignment.
Thus, NRGA++ can cope well with real-world datasets from depth sensors which do not provide hand shape annotations.  
Compared to the shape alignment counterparts from \cite{HandVoxNet2020}, GCN-MeshReg significantly improves in hand shape alignment accuracy over DispVoxNet, and NRGA++ improves both in runtime and accuracy over pointwise NRGA. 

\subsubsection{GCN-MeshReg: Neural Shape Registration} \label{sssec:DeepReg} 

In this subsection, we adopt a CNN-based method and propose a new GCN-based algorithm for shape registration, \textit{i.e.,}
DispVoxNets and GCN-MeshReg, respectively. 
Both these components enable end-to-end deep network for reconstructing a full 3D mesh and pose of a human hand from a single depth image. 

\noindent\textbf{CNN-based Shape Registration.} 
The original DispVoxNets~\cite{ShimadaDispVoxNets2019} is comprised of 
two stages, \textit{i.e.,}~global displacement estimation and refinement. 
The refinement stage removes roughness on the underlying surfaces represented as point sets. 
In contrast to the original approach \cite{ShimadaDispVoxNets2019}, we replace the refinement stage with Laplacian smoothing  \cite{vollmer1999improved}, as the hand mesh topology is known. 

In DispVoxNet, the hand surface shape $\hat{\mathcal{V}}_T$ is first converted into a voxelized grid $\hat{\mathcal{V}}^{'}_T$ of dimensions $64\times64\times64$. 
DispVoxNet estimates per-voxel displacements of the dimension $64^3\times3$ between the reference $\hat{\mathcal{V}}_S$ and voxelized hand surface $\hat{\mathcal{V}}^{'}_T$. 
The displacement loss $\mathcal{L}_{\textrm{Disp}}$ is given by: 
\begin{equation} 
\mathcal{L}_{Disp.}=\frac{1}{Q^{3}}\left\Vert \mathbf{d} - D_{vn}(\hat{\mathcal{V}}_S, \hat{\mathcal{V}}^{'}_T)\right\Vert^{2},
\end{equation}
where {$D_{vn}(\cdot)$ represents DispVoxNet based registration network from \cite{HandVoxNet2020}.} $Q$ and $\mathbf{d}$ are the voxel grid size and the ground-truth displacement, respectively. 
Since it is difficult to obtain $\mathbf{d}$ between the voxelized shape $\hat{\mathcal{V}}_S$ and hand surface $\hat{\mathcal{V}}_T$, the displacements are first computed between $\mathcal{V}_T$ and $\hat{\mathcal{V}}_T$, and are discretized to obtain $\mathbf{d}$. 
For more details on the computation of ground-truth voxelized grids, please refer to \cite{ShimadaDispVoxNets2019}. 

\noindent\textbf{GCN-based Shape Registration.} 
We propose in this article the first -- to the best of our knowledge -- mesh-to-voxel shape registration algorithm called {GCN-MeshReg}, which is based on GCNs. 
Recently, GCNs have enjoyed great attention in computer vision \cite{kolotouros2019cmr,ranjan2018generating,cheng2020faster}, and we find that they are a better alternative for such alignment problems compared to the CNN-based shape registration due to several reasons. 
First, GCNs have better ability to learn correct representations of graph-structured data. 
Second, the hand mesh registration problem can be addressed using  graph-based techniques as meshes can be naturally converted to graphs. 
Third, graph convolutions can learn inter-vertex relationships which lead to feasible and smooth hand meshes. 

Our GCN-MeshReg is inspired by \cite{kipf2016semi, kolotouros2019cmr} which use graph convolutions for data classification \cite{kipf2016semi} and single-image shape regression \cite{ kolotouros2019cmr}. 
GCN-MeshReg is defined for shape alignment and differs from \cite{kipf2016semi, kolotouros2019cmr} in several ways, see Table \ref{tab:dvn_architecture} for the architecture details of GCN-MeshReg. 
Our approach assumes that the reference shape is represented by voxel occupancy probabilities, and the template is a hand mesh. 
Moreover, it includes an iterative refinement cycle. 
GCN-MeshReg explicitly takes into account the mesh topology, 
unlike most of the algorithms which operate on point clouds and directly estimate the vertex positions or displacements \cite{ShimadaDispVoxNets2019,NRICP2003,jian2005robust}. 
By comparing the performance of the previous CNN-based registration approach DispVoxNets \cite{ShimadaDispVoxNets2019} with the proposed GCN-based alternative, \textit{i.e.,} GCN-MeshReg, we show that graph convolutions significantly increase the performance of the hand mesh registration, which we discuss in Sec.~\ref{sec:experiments}. 

GCN-MeshReg consists of two stages, \textit{i.e.,} feature extraction from the voxelized shape $\hat{\mathcal{V}}_S$ using the 3D convolution-based voxel feature extractor (VFE) and the shape registration stage $G_{cnn}$ using GCNs. 
In contrast to DispVoxNets, the graph convolutions in GCN-MeshReg utilize vertex connectivity information by constructing the row-normalized adjacency matrix  $\mathbf{\hat{A}}\in\mathbb{R}^{\textrm{K} \times \textrm{K}}$. 

See Fig.~\ref{fig:Pipeline}-(Stage 3) for an overview of GCN-MeshReg. 
First, VFE accepts $\hat{\mathcal{V}}_S$ and obtains its feature vector $\mathbf{f} \in\mathbb{R}^{2187}$. $\mathbf{f}$ is duplicated and concatenated on each vertex $ v  \in\mathbb{R}^{u}$ of $\hat{\mathcal{V}}_T$. 
The vertices in our dataset contain $x$-, $y$- and $z$-coordinates, hence $u=3$. 
$G_{cnn}$ accepts these vertices with the concatenated features along with $\mathbf{\hat{A}}$, applies graph convolutions and returns the registered output shape. 
We further repeat the same procedure for registrations of higher accuracy in the refinement cycle, using the output shape of the previous registration step as a new template shape input for $G_{cnn}$. %
After the refinement cycle, GCN-MeshReg returns the final 3D hand shape $\mathcal{V}_{out}$. 
Significance of the refinement cycle is highlighted in Table \ref{tab:HANDS19Shape} in Sec.~\ref{sec:experiments}, where we summarise the effect of our registration components on the final outcome. 

\begin{table}
\center
\resizebox{.99\linewidth}{!}{
\begin{tabular}{|c|c|c|c|c| }\hline
 \textbf{ID} & \textbf{Layer} & \textbf{Output Sz} & \textbf{Kernel Sz} & \textbf{Stride/Padding}  \\\hline
 1  &     Input & (1) 64x64x64 & - & -/-   \\\hline
 2  &     3D Conv & (16) 64x64x64 & 7x7x7 & 1/3  \\\hline %
 3  &     LeakyReLU & (16) 64x64x64 & - & -/-   \\\hline
 4  &     3D MaxPooling  & (16) 32x32x32 & 2x2x2 & 2/0  \\\hline
 
 5  &     3D Conv & (8) 32x32x32 & 5x5x5 & 1/2  \\\hline %
 6  &     LeakyReLU & (8) 32x32x32 & - & -  \\\hline
 7  &     3D MaxPooling & (8) 16x16x16 & 2x2x2 & 2/0  \\\hline
 
 8  &     3D Conv & (4) 16x16x16 & 3x3x3 & 1/1 \\\hline %
 9  &     LeakyReLU & (4) 16x16x16 & - & -/- \\\hline
 10 &   3D  MaxPooling  & (4) 8x8x8 & 2x2x2 & 2/0  \\\hline
 
 11 &     3D Deconv & (3) 9x9x9 & 4x4x4 & 1/1 \\\hline 
  12 &     Flatten & 2187 & - & - \\\hline 
 \end{tabular}
 }
 \vspace{+2mm}
\caption{\textbf{The architecture details of the voxel feature extractor (VFE) in GCN-MeshReg.} The negative slope for LeakyReLU is $0.01$.} 
\vspace{-6mm}
\label{tab:dvn_architecture}
\end{table}

For $G_{cnn}$, 
we follow the formulation similar to \cite{kipf2016semi}: 
\begin{equation} 
\mathbf{Y} = \sigma(\mathbf{\hat{A}}\mathbf{X}\mathbf{W}),
\end{equation}
where $\sigma(\cdot)$ denotes an activation function, 
$\mathbf{X}\in\mathbb{R}^{\textrm{K} \times \textrm{p}}$ is the input to the graph convolutional layer and $\mathbf{W}\in\mathbb{R}^{\textrm{p} \times \textrm{s}}$ is its weight matrix of the graph convolutional layer in $G_{cnn}$. $\textrm{p}$ denotes the dimensionality of the feature of each hand vertex. In our case, $\mathbf{X}$ of the first graph convolution layer is the template hand vertices with the concatenated $f$. $\mathbf{X}$ for the subsequent graph convolution layers are the output from the previous layers.
In this formulation, the row-normalized adjacency matrix  $\mathbf{\hat{A}}$ explicitly disconnects the feature aggregation from non-connected vertices, which helps the network to gather %
relevant features associated with the topology of the mesh data. 
$G_{cnn}$ operations with $i$ refinement steps are formulated as follows:
\begin{equation}\label{eq:final_output}
\mathcal{V}_{out} =  \left\{\begin{matrix}
G^{i=0}_{cnn}(\mathbf{\hat{A}},\mathbf{f},\hat{\mathcal{V}}_T)\;\;\;\;\;\;\;\;\;\;\;\;\;\;\;\;\;\;\;\;\;\;\;\\ \\ 
G^{i=n}_{cnn}(\mathbf{\hat{A}},\mathbf{f},G^{i=n-1}_{cnn}) \;\;\;(n > 0). 
\end{matrix}\right. 
\end{equation} 
The final output $\mathcal{V}_{out}$ is further integrated in the objective function $\mathcal{L}_{reg.}$:  
 
\begin{equation}\label{eq:graphConv} 
\mathcal{L}_{reg.}=\frac{1}{\textrm{K}}\left\Vert  \mathcal{V}_T- \mathcal{V}_{out}\right\Vert^{2}.
\end{equation} 

\subsubsection{Our Classical Shape Registration Algorithm} 
\label{sssec:GenReg} 
\begin{figure}[htbp]\centering
 \includegraphics[width=0.98\linewidth]{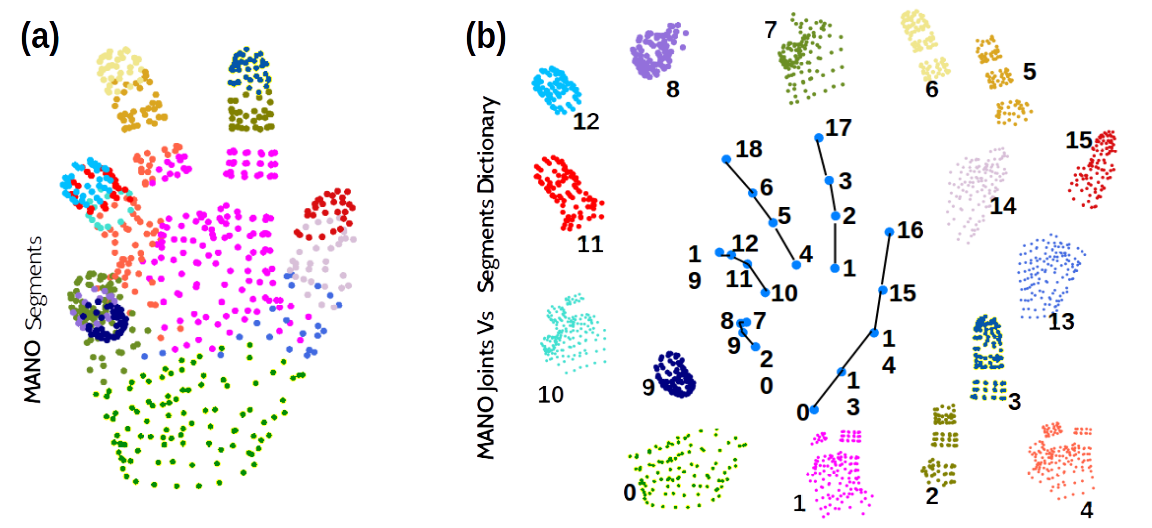}
\vspace{-3mm}
 \caption{\textbf{Selection of Segments in NRGA++}. 
 (a:) Pre-defined MANO hand shape segments in different colors. (b:) MANO hand model provides a prior mapping between 21 joints (except end-effectors as fingertips) and overlapping hand segments.} 
 \label{fig:MANODict} 
 \vspace{-6mm}
\end{figure}

Although neural registration methods achieve high accuracy in low runtime, these methods are bounded to the availability of training datasets. {Recent datasets such as HANDS19~\cite{armagan2020measuring} and HO-3D~\cite{hampali2020honnotate} are annotated for both the hand pose and shape. However, several benchmarks (\textit{e.g.,} MSRA~\cite{sun2015cascaded}, NYU~\cite{tompson2014real}, ICVL~\cite{tang2014latent}, 
MSRC~\cite{sharp2015accurate},
FHAD~\cite{garcia2018first} \textit{etc.}) do not provide hand shape annotations. Therefore, if the ground truth of hand shape is not available, we need an alternative to a neural registration method.}
For this reason, alternatively to DispVoxNet and GCN-MeshReg (Sec.~\ref{sssec:DeepReg}), we propose NRGA++, \textit{i.e.,} a classical physics-based algorithm for registering the template mesh $\hat{\mathcal{V}}_T$ with the reference voxelized hand $\hat{\mathcal{V}}_S$, which is a modified version of NRGA~\cite{Ali_NRGA_2018}. 
Our choice falls to NRGA as it preserves local hand mesh  topology and is robust to noise in $\hat{\mathcal{V}}_S$. 
NRGA splits the template point sets into overlapping regions, each of which is registered rigidly to the corresponding region of influence in the reference. 
The final per-point displacement is obtained as a consensus rigid transformation among the segments in which a given point is involved.
NRGA is a computationally-expensive iterative method, and the updates of point positions are performed by simulating Newtonian particle dynamics. 
In NRGA++, we thus avoid the general-purpose automatic region allocation step and select the template regions as segments of the MANO hand model, as visualized in Fig.~\ref{fig:MANODict}. 
Moreover, Fig.~\ref{fig:MANODict} visualises how $21$ joints of the MANO model are mapped to the overlapping hand segments. 
For the end-effectors, \textit{e.g.,} we select half of the vertices present in the segment nearest joint. 
Apart from that, all steps of NRGA++ are as in the original NRGA (\textit{e.g.,} $k$-d tree building and calculating the consensus per-point transformations), see \cite{Ali_NRGA_2018} for more details. 
{The proposed segment-wise point set alignment strategy leads to a significant improvement in runtime performance of NRGA++. Moreover, NRGA++ relies on the topological information \textit{i.e.,} vertex connectivity graph; thereby, it preserves the structure of the hand mesh.}

\vspace{0.25cm}
\subsection{Data Augmentation in 3D} 
\label{ssec:DataAugmentation} 
Our method for hand shape estimation relies on the accuracy of the estimated 3D pose. 
Therefore, the hand pose estimation method has to be accurate and robust.  
Training data augmentation helps to improve the performance of a deep network \cite{oberweger2017deepprior++}. 
Existing methods for hand pose estimation~\cite{oberweger2017deepprior++,tompson2014real} use data augmentation in 2D. 
This is mainly because these methods treat depth maps as 2D data. 
The representation of the depth map in voxelized form makes it convenient to perform data augmentation in all three dimensions. In this paper, we perform 3D data augmentation which improves the accuracy and robustness of hand pose estimation (see Sec.~\ref{ssec:EvalHandPose}). 

\begin{figure*}[t]
\begin{center}
      \includegraphics[width=0.99\linewidth]{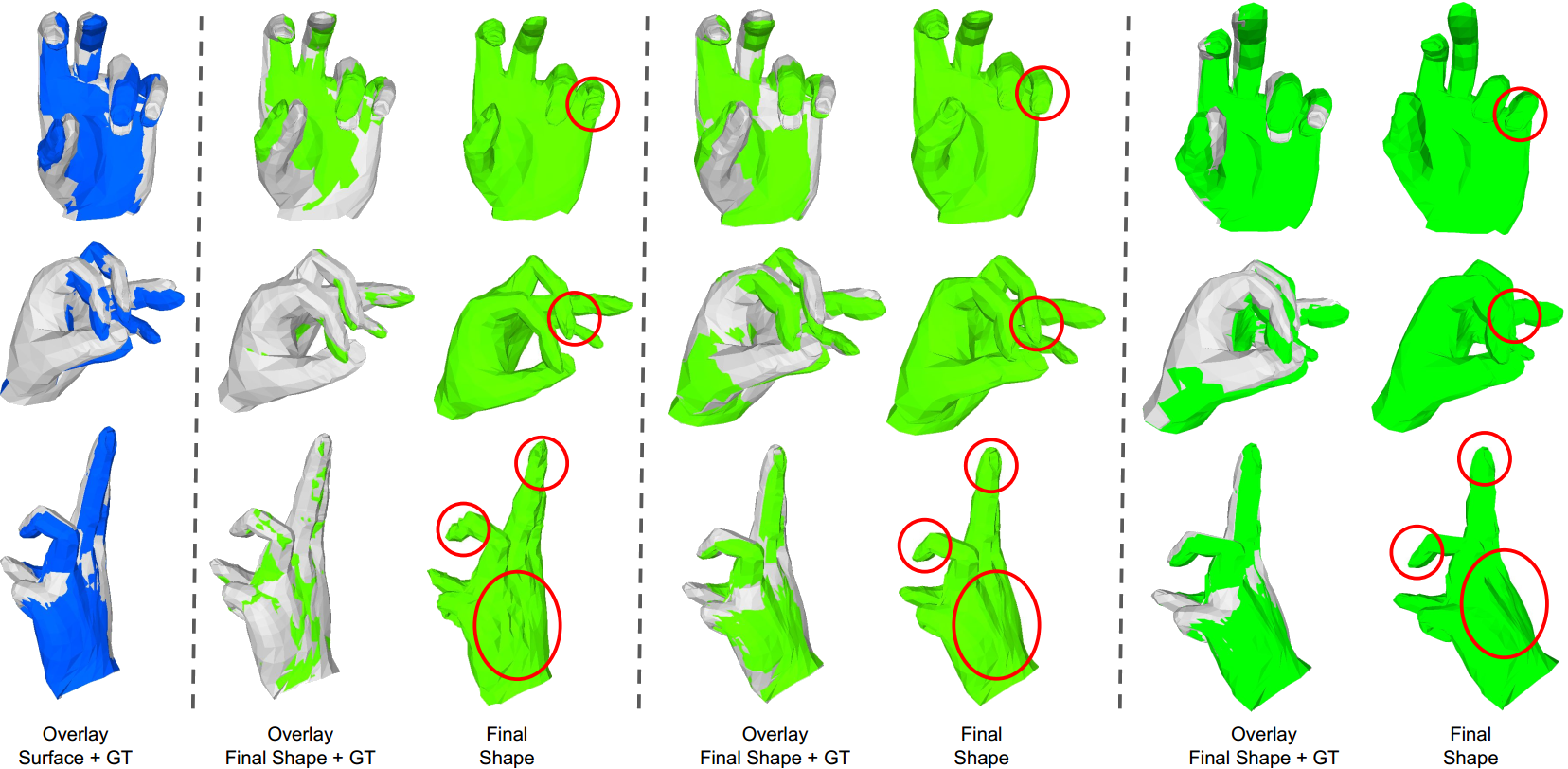}
      
      \rotatebox{0}{\hspace{0mm}(a) w/o regis. \hspace{20mm}(b) DispVoxNet \hspace{25mm} (c) \textbf{GCN-MeshReg}\hspace{15mm} \hspace{20mm} (d) NRGA\hspace{15mm}}
\end{center}
   \vspace{-4mm}
   \caption{\textbf{Qualitative
   comparison of different registration methods on  SynHand5M~\cite{malik2018deephps} dataset.} 
   For a fair comparison, we use the HandVoxNet~\cite{HandVoxNet2020} framework and replace only the registration component by: (a) no registration; see the estimated hand surface (blue) and the ground truth (grey). 
   (b) DispVoxNet; see artefacts and roughness in the estimated  shapes. (c) GCN-MeshReg produces much smoother and more  accurate final shapes. (d) NRGA produces smoother hand  shapes, but the registered shapes are visually less accurate  and slightly blown up. 
   }
\label{fig:Synregcomp}
\vspace{-2mm}
\end{figure*}

During V2V-PoseNet training, we apply simultaneous rotations in all three axes (\textrm{x},\textrm{y},\textrm{z}) to each 3D coordinate ($i,j,k$) of $\textrm{V}_\textrm{D}$ and $\mathcal{H}_j$ by using Euler transformations: 
\begin{equation} \label{eq:augmentation}
[\hat{i},\hat{j},\hat{k}]^\textrm{T} = 
[\textrm {Rot}_\textrm{x}(\theta_\textrm{x})]\times[\textrm {Rot}_\textrm{y}(\theta_\textrm{y})]\times
[\textrm {Rot}_\textrm{z}(\theta_\textrm{z})][i,j,k]^\textrm{T}, 
\end{equation}
where ($\hat{i},\hat{j},\hat{k}$) is the transformed voxel coordinate. 
$\textrm {Rot}_\textrm{x}(\theta_\textrm{x})$, $\textrm {Rot}_\textrm{y}(\theta_\textrm{y})$ and $\textrm {Rot}_\textrm{z}(\theta_\textrm{z})$ are $3\times3$ rotation matrices around $\textrm{x}$, $\textrm{y}$ and $\textrm{z}$ axes. 
The values for $\theta_\textrm{x}$, $\theta_\textrm{y}$ and $\theta_\textrm{z}$ are selected randomly in the ranges $[$$-40^{\circ}$, +$40^{\circ}$$]$, $[$$-40^{\circ}$, +$40^{\circ}$$]$ and $[$$-120^{\circ}$, +$120^{\circ}$$]$, respectively. 
In addition to rotations in 3D, following \cite{moon2017v2v}, we perform scaling and translation in the respective ranges $[+0.8, +1.2]$ and $[-8,+8]$. 

\vspace{7pt} 

\section{The Network Training} 
\label{sec:NetTraining}
 The network input $\textrm{V}_\textrm{D}$ is generated by projecting the raw depth image pixels into 3D space. Hand region points are then extracted by using a cube of size $250$ that is centered on hand palm center position. 
3D point coordinates of the hand region are discretized in the range $[$$1$, $88$$]$. 
Finally, to obtain $\textrm{V}_\textrm{D}$, each voxel is assigned a value using TSDF (see Eq.~\ref{eq:tsdf}). 
Similarly, $\mathcal{V}_S$ is obtained by voxelizing the hand mesh except the final step where the voxel value is set to $1$ for the 3D point coordinate of the hand region, and $0$ otherwise. 
Following\mbox{~\cite{moon2017v2v}}, $\mathcal{H}_j$ are generated as 3D Gaussians.
${\mathcal{V}}_T$  is created by normalizing the mesh vertices in the range $[$$-1$, $+1$$]$. 
We perform this normalization by subtracting the mesh vertices from the hand palm center and then dividing them by half of the cube size. 
We train our V2V-PoseNet 
on the datasets described in Sec.\mbox{~\ref{sec:experiments}} 
using the 3D data augmentation technique mentioned in Sec.\mbox{~\ref{ssec:DataAugmentation}}.
For all the datasets, we use the learning rate (LR) of $2.5e^{-4}$, batch size = $8$, and RMSProp as an optimization method.  
We train V2S-Net and V2V-ShapeNet independently on each of the datasets using $2.5e^{-4}$ and $0.5e^{-3}$ as learning rates, respectively. The value of batch size is $8$ and RMSProp is used for the optimization of both the shape networks. 
DispVoxNet and GCN-MeshReg are trained using Adam optimizer\mbox{~\cite{adam}} with learning rate of $3.0e^{-4}$. 
The training continues until the convergence of $\mathcal{L}_{\textrm{Disp}}$ and  $\mathcal{L}_{\textrm{reg}}$ with batch size $12$, respectively. 
We first train VFE and $G_{cnn}$ for the registration using the ground-truth hand shapes. 
Then, we train another instance of $G_{cnn}$ for the refinement step using the output from the registration step as inputs to the $G_{cnn}$. All models are trained until convergence on a desktop workstation equipped with Nvidia Titan X GPU. 

 \begin{table}[t]
\begin{center}
\begin{tabular}{|l|c|c|}
\hline
\textbf{Methods} & \textbf{V2V-ShapeNet} & \textbf{V2S-Net}\\
         &  \textbf{3D} $\boldsymbol{\mathcal{S}}$ \textbf{Err.}  &  \textbf{3D} $\boldsymbol{\mathcal{V}}$ \textbf{Err. (\textit{mm})}\\
\hline\hline
w/o $\mathcal{H}_j$ & 0.007 & 8.78 \\
w/o ${\textrm{V}}'_\textrm{D}$ & 0.016 & 3.54 \\
with ($\mathcal{H}_j\oplus{\textrm{V}}'_\textrm{D}$) & \textbf{0.005} & \textbf{3.36} \\
\hline
\end{tabular}
\end{center}
\caption{\textbf{Ablation study on inputs (\textit{i.e.,}~$\mathcal{H}_j$ and ${\textrm{V}}'_\textrm{D}$) to V2S-Net and V2V-ShapeNet on SynHand5M~\cite{malik2018deephps}.} 
We observe that combining both inputs is useful for these two networks. 
}
\label{tab:AblationStudyConcatenation}
\vspace{-6mm}
\end{table}

\begin{table}[t]
\begin{center}
\begin{tabular}{|l|c|}
\hline
\textbf{Methods} & \textbf{3D} $\boldsymbol{\mathcal{V}}$ \textbf{Err. (\textit{mm})} \\
\hline\hline
DeepHPS \cite{malik2018deephps} & 11.8 \\
WHSP-Net \cite{malik2019whsp} & 5.12 \\
HandVoxNet \cite{HandVoxNet2020} (with DispVoxNet) &  2.92  \\
HandVoxNet \cite{HandVoxNet2020} (with GCN-MeshReg w/o ref.) &  \textbf{1.79}  \\
HandVoxNet \cite{HandVoxNet2020} (with GCN-MeshReg) &  \textbf{1.72}  \\
\hline
\end{tabular}
\end{center}
\caption{{\textbf{Comparison of different registrations and the state of the arts on SynHand5M~\mbox{\cite{malik2018deephps}}.} 
Notably, for a fair comparison, we replace only the registration component of~\cite{HandVoxNet2020}, and keep all other components same. Our graph convolutions based registration with refinement outperforms 3D-convolution-based method (\textit{i.e.,} DispVoxNet) by $41.09$\%.} 
}
\label{tab:ShapeResultsSynthetic}
\vspace{-7mm}
\end{table}

\begin{figure*}[t]
\begin{center}
      \includegraphics[width=0.99\linewidth]{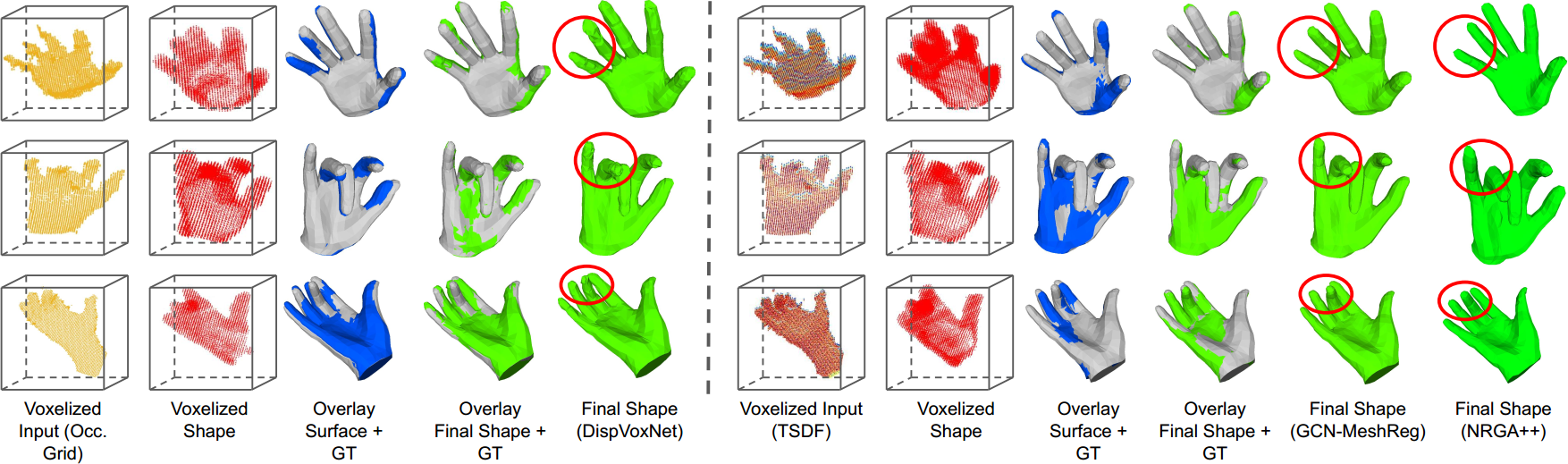}      
   
   \rotatebox{0}{\hspace{1mm}(a) HandVoxNet~\cite{HandVoxNet2020} \hspace{70mm} (b) Our Method}
\end{center}
   \vspace{-4mm}
   \caption{\textbf{Qualitative
   Comparison of  3D hand shape estimation on HANDS19 \cite{armagan2020measuring} dataset:} We present a one-to-one comparison of the intermediate and final predictions of the two compared configurations. The first column shows the 3D voxelized depth maps. The second column shows the estimated 3D voxelized hand shapes. The third column shows the overlay of the  estimated and the ground-truth hand surfaces. The fourth column shows the overlay of the registered (using the convolutional networks) and the ground-truth hand surfaces. 
   The fifth column shows the final shapes. 
   The right-most column (``Our Method") shows the registered hand shapes using the NRGA++. 
   The red circles highlight the regions where significant  differences in the estimations can be observed. 
   Our approach with graph-convolution-based registration (\textit{i.e.,} GCN-MeshReg) shows visually smoother and accurate final shapes.  
   }
\label{fig:OursVsHandVoxNet}
\vspace{-2mm}
\end{figure*}

\section{Experiments} 
\label{sec:experiments} 
{We perform qualitative and quantitative evaluations of our approach on four challenging hand pose datasets, namely, 
HO-3D\mbox{~\cite{hampali2020honnotate}}, HANDS19  Challenge (Task 1)\mbox{~\cite{armagan2020measuring}}, 
BigHand2.2M (HANDS17  Challenge)\mbox{~\cite{yuan2017bighand2}} and 
SynHand5M\mbox{~\cite{malik2018deephps}} datasets. 
An ablation study on the inputs of the proposed shape  estimation network is performed on  SynHand5M\mbox{~\cite{malik2018deephps}} dataset. 
}

\subsection{Datasets and Evaluation Metrics}
\label{ssec:metrics}
{HO-3D\mbox{~\cite{hampali2020honnotate}} and HANDS19 Challenge\mbox{~\cite{armagan2020measuring}} are the recent real benchmarks that provide annotations for 3D hand pose- and shape-based on the MANO model\mbox{~\cite{romero2017embodied}}. Task $1$ of HANDS19 Challenge is depth-based hand pose estimation which builds on BigHand2.2M\mbox{~\cite{yuan2017bighand2}}. It contains hands in isolation from both egocentric and third-person viewpoints.
The dataset provides $175,951$ training and $124,999$ testing frames. 
For evaluation, 3D hand pose estimations of the test set are submitted to an online portal\footnote{\url{https://competitions.codalab.org/competitions/20913}} which shows the achieved accuracy on the leader-board.
HO-3D dataset provides annotated RGB-D frames of hands manipulating with the objects. The training data consists of $66,034$ frames and the test set is made of $11,524$ images. The dataset contains $68$ sequences which are recorded from ten different subjects manipulating with ten distinct objects. The predictions of both 3D hand pose and shape are uploaded in a specified format on web portal\footnote{\url{https://competitions.codalab.org/competitions/22485?}} for obtaining the accuracies, and comparisons with other submissions can be seen on the portal.
SynHand5M\mbox{~\cite{malik2018deephps}} is the largest synthetic dataset that contains fully annotated five million depth images for both the 3D hand pose and shape. The sizes of its training ($\mathcal{T}_\textrm{S}$) and test sets are $4.5$\textrm{M} and $500$\textrm{k}, respectively. The hand model of SynHand5M is created synthetically\mbox{~\cite{malik2018deephps}}, which is different from the MANO model.
BigHand2.2M\mbox{~\cite{yuan2017bighand2}} is a million-scale real hand pose dataset. It does not provide hand shape annotations. For pose estimation, it provides accurate joint annotations for $956$\textrm{k} training depth images acquired from ten subjects. 
The size of the BigHand2.2M's test set is $296$\textrm{k}. 
The results of 3D hand pose estimation are submitted to HANDS17 Challenge\footnote{\url{https://bit.ly/3iFmlzf}}, which provides quantitative comparisons with the other submissions.  

We use three evaluation metrics: (i) the average 3D joint location error over all test frames (3D $\mathcal{J}$ Err.); (ii) mean vertex location error over all test frames (3D $\mathcal{V}$ Err.); and (iii) mean voxelized shape error (\textit{i.e.,}~per-voxel binary cross-entropy) over all test data (3D $\mathcal{S}$  Err.). 
}

\subsection{Evaluation of Hand Shape Estimation}
\label{ssec:EvalShapeEstim}
In  this  subsection,  we demonstrate the ability of our algorithm to reconstruct real hand shapes. Moreover, we illustrate the effectiveness of our design choice by conducting an ablation study on the inputs (\textit{i.e.,}~$\mathcal{H}_j$ and ${\textrm{V}}'_\textrm{D}$). 
To this end, we  evaluate  our  proposed  approach  on  SynHand5M, HANDS19 Challenge (Task 1) and HO-3D datasets.
  
{\noindent\textbf{Synthetic Hand Shape Reconstruction.}} 
We train 
our HandVoxNet pipeline %
on SynHand5M by following the training method described in \mbox{Sec.~\ref{sec:NetTraining}.} 
We conduct an ablation study on the inputs (\textit{i.e.,} ${\textrm{V}}'_\textrm{D}$ and $\mathcal{H}_j$) of V2V-ShapeNet and V2S-Net to show the effectiveness of our design choice.
We regress  $\hat{\mathcal{V}}_T$ and $\hat{\mathcal{V}}_S$ by 
using input ${\textrm{V}}'_\textrm{D}$ (\textit{i.e.,}~without  $\mathcal{H}_j$).
Similar experiments are repeated by providing $\mathcal{H}_j$ (\textit{i.e.,}~without ${\textrm{V}}'_\textrm{D}$) and $\mathcal{I}_S$ (\textit{i.e.,}~with $\mathcal{H}_j \oplus {\textrm{V}}'_\textrm{D}$) as separate inputs to V2V-ShapeNet and V2S-Net.
The results are summarized in \mbox{Table~\ref{tab:AblationStudyConcatenation}} that clearly show the advantage of concatenating voxelized depth map with 3D heatmaps of joints. 

\mbox{Table~\ref{tab:ShapeResultsSynthetic}} compares our algorithm (with different neural registration approaches) and  state-of-the-art hand shape estimation algorithms.
DispVoxNet~\mbox{\cite{HandVoxNet2020}} fits $\hat{\mathcal{V}}_T$ to the $\hat{\mathcal{V}}_S$, which results in $13.1$\% improvement in the surface reconstruction.
Despite this quantitative improvement in the accuracy, the final shape suffers from severe artefacts, as shown in \mbox{Fig.~\ref{fig:Synregcomp}(b)}.
Our proposed GCN-MeshReg not only outperforms DispVoxNet by $35.6$\% but also effectively removes artefacts in the shape estimation 
(see \mbox{Table~\ref{tab:ShapeResultsSynthetic}} and \mbox{Fig.~\ref{fig:Synregcomp}(c)}).
{For a fair comparison of the above two registration methods, we use the occupancy grid representation \mbox{\cite{moon2017v2v,HandVoxNet2020}} of voxelized depth map as input to our approach.}  
Furthermore, the accuracy estimated $\mathcal{V}_S$ (\mbox{Table  \ref{tab:AblationStudyConcatenation}}) is higher compared to WHSP-Net (\mbox{Table~\ref{tab:ShapeResultsSynthetic}}), which clearly shows the effectiveness of employing 3D-CNN-based network for direct mesh vertex regression.  

\begin{table}[t]
\begin{center}
\scalebox{0.88}{
\begin{tabular}{|l|c|c|}
\hline
 
\textbf{Method}  & \textbf{3D} $\boldsymbol{\mathcal{V}}$ \textbf{Err. (\textit{mm})} &  \textbf{Err. reduction} \\
         &  \textbf{With Reg.}  &      \textbf{after the Reg. }$\boldsymbol{(\%)}$\\
\hline\hline
HandVoxNet \cite{HandVoxNet2020} (w/o Reg.)$\dagger$  & 4.71  & - \\
HandVoxNet \cite{HandVoxNet2020} (with DispVoxNet)  & 4.14  & 12.1 \\
\hline 
{Ours} (w/o Reg.)  &  4.90 & - \\
{Ours} (with GCN-MeshReg w/o ref.) & 4.08   & 16.7  \\ %
{Ours} (with GCN-MeshReg) & \textbf{3.57}  & \textbf{27.1}  \\ %
{Ours} (with NRGA++) & 4.5  & 8.2  \\
\hline 
\end{tabular}}
\end{center}
\caption{\textbf{3D hand shape estimation results on HANDS19 (Task 1) \cite{armagan2020measuring} dataset.} 
The left column shows methods that use different registrations. Middle column shows the accuracy of the final shapes after the registration. 
The last column shows the percentage shape error reduction after registration.
We observe significant improvement with GCN-MeshReg.
The proposed registration method improves the shape estimation by $13.7$\% compared to HandVoxNet~\cite{HandVoxNet2020}. 
}
\label{tab:HANDS19Shape}
\vspace{-6mm}
\end{table}

\begin{figure}[t]
\begin{center}
  \includegraphics[width=1\linewidth]{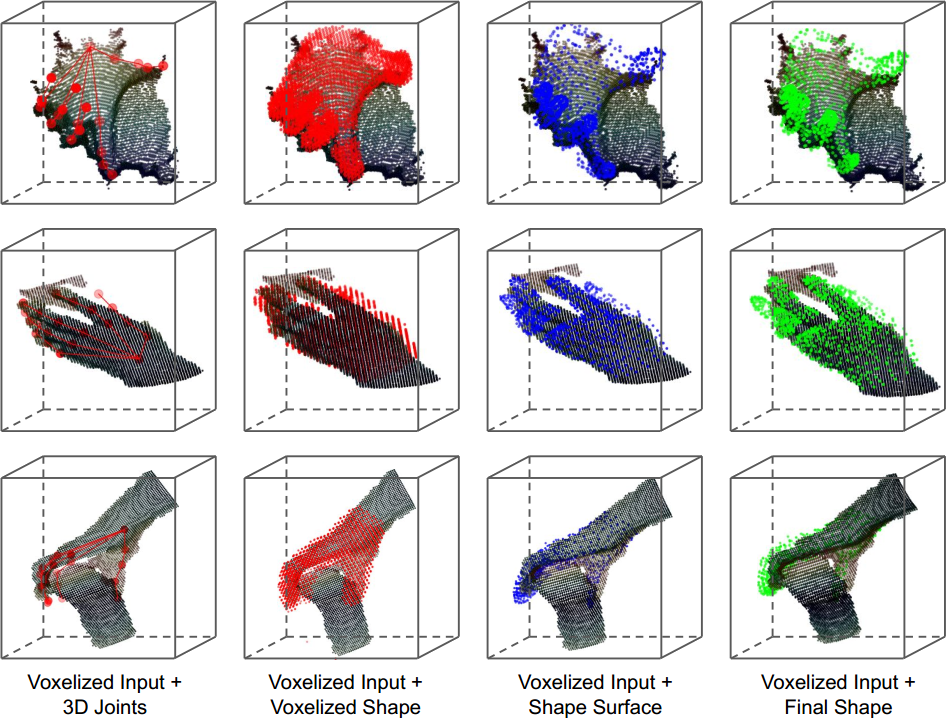}
\end{center}
   \vspace{-3mm}
  \caption{\textbf{Qualitative results on HO-3D \mbox{\cite{hampali2020honnotate}} dataset.}
  From left to right, we show overlays of the estimated 3D pose, 3D voxelized hand shape, hand surface and the final (registered) shape on the sample 3D voxelized depth inputs, respectively. 
  } 
\label{fig:ho3dQualitative}
\end{figure}

\begin{table}[t]
\begin{center}
\begin{tabular}{|l|c|c|c|c|c|}
\hline
\textbf{Methods} & \textbf{Main Err.} & \textbf{\textrm{I}.} & \textbf{\textrm{S}.} & \textbf{\textrm{A}.} & \textbf{\textrm{V}.} \\
\hline\hline
BT \cite{yang2019aligning} & 23.62 & 18.78 & 21.84 & 16.73 & 19.48 \\
IPR \cite{sun2018integral} & 19.63 & 8.42 & 14.21 & 7.5 & 14.16 \\
NTIS \cite{moon2017v2v} & 15.57 & 4.54 & 12.05 & 4.21 & 8.47 \\
AWR \cite{huang2020awr} & 13.76 & \textbf{3.93} & 11.75 & \textbf{3.65} & \textbf{7.50} \\
A2J \cite{xiong2019a2j} & 13.74 & 6.33 & 11.23 & 6.05 & 8.78 \\
\hline
V2V\cite{moon2017v2v} & 16.64 & 5.46 & 11.86 & 5.35 & 8.96 \\
HandVoxNet\cite{HandVoxNet2020} & 15.57 & 5.98 & 11.48 & 5.73 & 9.12 \\
\textbf{Ours} (Proj. D-TSDF) & 13.94 & 6.18 & \textbf{10.53} & 5.89 & 8.98 \\
\textbf{Ours} (Proj. TSDF) & \textbf{13.35} & 6.08 & 10.69 & 5.83 & 8.91 \\
\hline
\end{tabular}
\end{center}
\caption{\textbf{Comparison with the state-of-the-art methods in the task of depth-based 3D pose estimation  in  HANDS  2019  challenge.} 
Our projective-TSDF-based implementation  outperforms several methods in the main error and ranks first in the challenge. These error metrics are provided in $mm$. 
}
\label{tab:Hands19Quantitative}
\vspace{-7mm}
\end{table}

\begin{table}[t]
\begin{center}
\begin{tabular}{|l|l|c|}
\hline
\textbf{Methods} & \textbf{Components} & \textbf{Runtime, \textit{sec.}} \\
\hline\hline
& V2V-PoseNet &  0.011 \\
& V2V-ShapeNet & 0.0015  \\
HandVoxNet\cite{HandVoxNet2020} & V2S-Net &   0.0038\\
& DispVoxNet (GPU + CPU)$^*$ & 0.162 \\
& NRGA (CPU)    & 59 - 70 \\
\hline
& V2V-PoseNet (Proj. TSDF) & 0.022  \\
& V2V-ShapeNet (Proj. TSDF) & 0.0021  \\
HandVoxNet++ & V2S-Net (Proj. TSDF) & 0.0043  \\
& NRGA++    & 0.45  \\
& GCN-MeshReg   &  0.0341  \\
\hline
\end{tabular}
\end{center}
\caption{\textbf{Runtimes:} forward-pass of deep networks on GPU. ``$^*$'' shows that the Laplacian smoothing runs on CPU.
The transformation of a depth map to TSDF-based representation is more time consuming compared to the occupancy grid. The execution times of NRGA++ and GCN-MeshReg  are significantly improved over NRGA and DispVoxNet, respectively. 
} 
\label{tab:Runtime} 
\vspace{-6mm}
\end{table}

{\noindent\textbf{Real Hand Shape Reconstruction.}} 
{HANDS19 Challenge dataset provides the shape annotations only for its training set. Therefore, to evaluate the performance of our complete method configuration on this dataset}, we select approximately $90$\% of the original training data (\textit{i.e.,} $158,355$ frames) as the new train set ($\mathcal{T}_\textrm{H}$) and the remaining data (\textit{i.e.,} $17595$ frames) as our test set.
We train V2S-Net and V2V-ShapeNet on $\mathcal{T}_\textrm{H}$ using the hyperparameters mentioned in \mbox{Sec.~\ref{sec:NetTraining}}. 
V2S-Net and V2V-ShapeNet accurately recover the hand shape representations. It is observed that the voxelized shape is more accurately estimated than the hand surface. Thereby, the alignment further refines the hand surface.
{\mbox{Table~\ref{tab:HANDS19Shape}} summarizes the quantitative comparison of the proposed HandVoxNet++ with HandVoxNet\cite{HandVoxNet2020}.
We train the method proposed in \mbox{\cite{HandVoxNet2020}} on the HANDS19.} We observe that the error reduction after the proposed GCN-MeshReg is $27.1$\%, while $12$\% error reduction is achieved for DispVoxNet-based registration \mbox{\cite{HandVoxNet2020}}.
A detailed qualitative comparison of the estimated shape representations is shown in \mbox{Fig.~\ref{fig:OursVsHandVoxNet}}. Notably, the artefacts can be clearly seen in the final shape estimation from DispVoxNet whereas, our proposed GCN-MeshReg does not suffer from such limitations and produces visually more accurate and smoother hand shapes thanks to the graph-convolution-based registration. 

HO-3D is a new and challenging hand pose and shape dataset with hands interacting with objects. We train HandVoxNet++ on this dataset using the hyperparameters described in \mbox{Sec.~\ref{sec:NetTraining}}. 
{The average 3D mesh error on the test set of HO-3D comes out to be $2.70$\textit{cm}.}
Because of high occlusion and appearance of random objects in the depth images, hand shape estimation task becomes harder. 
Despite we successfully recover plausible hand shapes on some challenging input voxelized depth maps (as shown qualitatively in \mbox{Fig.~\ref{fig:ho3dQualitative}}). 

{\noindent\textbf{{Runtime.}}} The runtimes of different components of HandVoxNet\mbox{\cite{HandVoxNet2020}} and our proposed HandVoxNet++ are summarized in \mbox{Table~\ref{tab:Runtime}}.
The proposed GCN-MeshReg is $78.9$\% faster than DispVoxNet. Also, NRGA++ is ${\sim}150$ times faster than NRGA on our hardware. With GCN-MeshReg, the runtime of each component is fast enough to develop a real-time interactive application when they are operated in parallel.
Although TSDF representation of depth map allows improving the accuracy of 3D pose estimation however, 
generating this representation is more time consuming than producing the occupancy gird representation. 

\begin{figure}[t]
\begin{center}
   \rotatebox{90}{\hspace{16mm} \small{Ours} \hspace{25mm} \small{HandVoxNet}~\cite{HandVoxNet2020}}
   \includegraphics[width=0.95\linewidth]{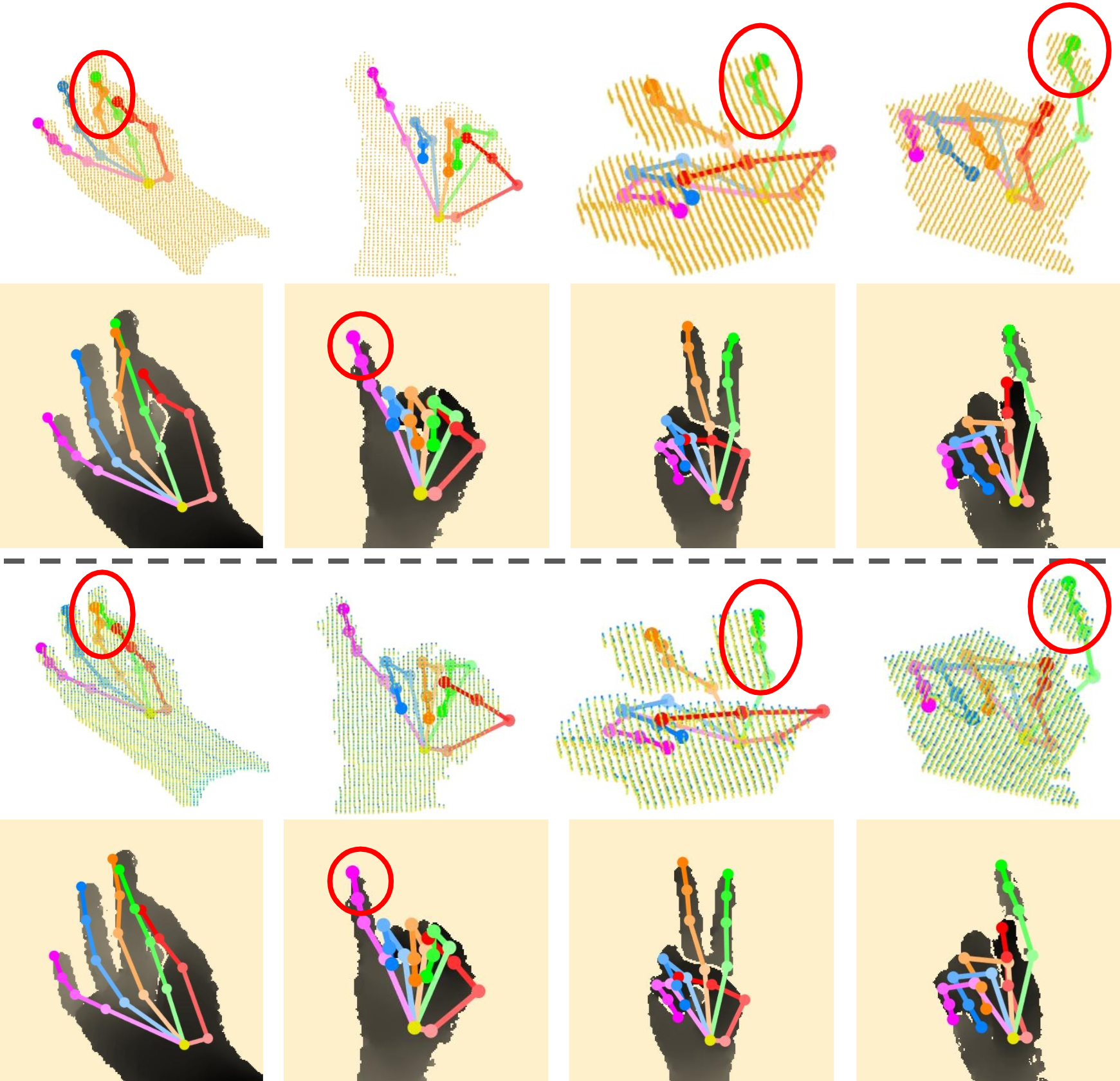}
\end{center}
   \vspace{-3mm}
   \caption{\textbf{
   Qualitative results of depth-based 3D pose estimation on HANDS19~\cite{armagan2020measuring} dataset.}
   The top row shows the overlay of the estimated 3D pose on 3D voxelized depth input, and the bottom row shows the corresponding 2D overlay on to the depth image. 
   The 3D view helps to better visualize the error in the estimated third dimension.
   Our 3D pose estimation results are more accurate compared to the occupancy-grid-based method~\cite{HandVoxNet2020}, see the red circles for their differences. 
   }
\label{fig:PoseQualitative}
\vspace{-4mm}
\end{figure}

\subsection{Evaluation of Hand Pose Estimation} 
\label{ssec:EvalHandPose} 
As it can be seen in HandVoxNet++ pipeline, the shape accuracy depends on the accuracy of the estimated pose. Therefore, 
the hand pose estimation needs to be robust and accurate. 
{In this work, we employ TSDF-based 3D voxelized representation which better encodes the 3D information of 2D depth map. We train V2V-PoseNet\mbox{\cite{moon2017v2v}} using this representation on HANDS19 Challenge dataset for 3D pose estimation from single depth images (Task 1).} 
This dataset contains hand images from egocentric and third-person viewpoints with no interaction with objects. 
The following error metrics are used for rigours evaluation; (i) \textbf{Main Err.}: It is the extrapolation error which uses test data containing hand shapes, viewpoints and poses that does not exist in the training set, (ii) \textbf{I.}: It is the interpolation error where test data contains hand shapes, poses and viewpoints which exist in the training set, (iii) \textbf{S.}: The test data contains hand shapes which are not present in the training data, (iv) \textbf{A.}: The test data contains hand poses which are not present in the training data, (v) \textbf{V.}: The test data contains viewpoints which are not included in the training set. 
Our results and comparison with the state-of-the-art methods are reported in \mbox{Table~\ref{tab:Hands19Quantitative}}. 
For a fair comparison, we also train the networks of \mbox{\cite{moon2017v2v}} and \mbox{\cite{HandVoxNet2020}} on HANDS19 Challenge dataset. Our method that uses the projective TSDF representation outperforms \mbox{\cite{moon2017v2v}} and \mbox{\cite{HandVoxNet2020}} by $19.8$\% and $14.2$\%, respectively. 
We observe that---in our case where a one-to-one mapping exists between the voxelized depth map and the 3D heatmaps---the projective TSDF representation converges faster and produces more accurate results compared to the projective directional TSDF representation.
The qualitative comparison of our method with HandVoxNet \mbox{\cite{HandVoxNet2020}} on some challenging hand poses is shown in \mbox{Fig.~\ref{fig:PoseQualitative}}.
Among the state-of-the-art methods, our approach is\textbf{ ranked first} during the moment of submission of our result on the web portal. Our method achieves $2.8$\% improvement in accuracy compared to Anchor-to-Joints (A2J) regression approach \mbox{\cite{xiong2019a2j}} which uses an ensemble predictions of $3$ epochs. Also, in comparison to other existing approaches such as Adaptive Weighting Regression (AWR)\mbox{\cite{huang2020awr}} with an ensemble of five epochs, NTIS\mbox{\cite{moon2017v2v}} with complex post-processing step including temporal smoothing, our method exceeds in the accuracy by $3.0$\% and $14.2$\%, respectively. 

We train our best performing pose estimation network (that uses projective representation) on HO-3D dataset. 
{By following \mbox{\cite{hampali2020honnotate}} and according to the web challenge, we report the pose error in $cm$. The mean 3D joint location error on the test set of HO-3D is $2.46$\textit{cm}.}
To the best of our knowledge, there is no published work so far which reports accuracy on this dataset using depth images. 
Qualitative results of pose estimation are shown in \mbox{Fig.~\ref{fig:ho3dQualitative}} (first column) on some challenging test samples where the hands are interacting with different objects.

Notably, our focus is to develop an effective approach for simultaneous hand pose and shape estimation. However, for completeness, we also train our method on {HANDS17 Challenge\footnote{\url{http://icvl.ee.ic.ac.uk/hands17/challenge/}} dataset which is created by sampling images from BigHand2.2M\mbox{\cite{yuan2017bighand2}} and FHAD\mbox{\cite{garcia2018first}} datasets}. The joint location error on the test set of HANDS17 dataset comes out to be $8.92$\textit{mm} which shows $10.3$\% and $3.7$\% improvement in accuracy compared to V2V\mbox{\cite{moon2017v2v}} (\textit{i.e.,} $9.95$\textit{mm}) and HandVoxNet\mbox{\cite{HandVoxNet2020}} (\textit{i.e.,} $9.27$\textit{mm}), respectively.

\begin{figure}[t]
\begin{center}
   \includegraphics[width=1.0\linewidth]{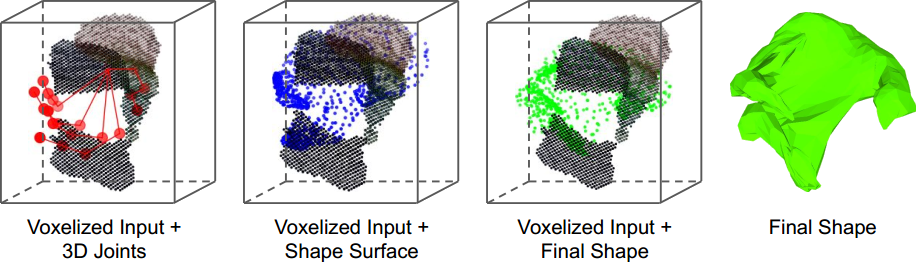}
\end{center}
   \vspace{-3mm}
   \caption{
   \textbf{Failure case.} Our method is unable to produce plausible shapes in cases of severe occlusion of hand parts. We show our pose and shape estimations on a challenging input where most hand parts are occluded by an object.    
   } 
   \label{fig:failure_dvn}
\vspace{-3mm}
\end{figure}

\section{Discussion} 
Our pipeline relies on the availability of the ground truth of real hand shapes thus, we do not employ weak supervision or training with combined real and synthetic data. 
For this reason, we do not experiment with the older datasets such as NYU~\cite{tompson2014real} and ICVL~\cite{tang2014latent} which do not provide annotations of hand shapes. Rather we experiment with the state-of-the-art datasets (\textit{i.e.,} HANDS19, HO-3D and SynHand5M) which provide annotations for both 3D hand pose and shape. 
We observe in the experiments that HandVoxNet++ significantly outperforms our previous method HandVoxNet~\cite{HandVoxNet2020} in the pose and shape accuracy, and runs faster. 
Our recovered hand shapes are smoother, and are less prone to artefacts. 
The reasons are manifold. 
First, it is due to the improved neural alignment component GCN-MeshReg with graph convolutional networks and iterative refinement cycle which incorporates the structure of hand mesh. 
Second, we found the projective TSDF representation establishes a more efficient one-to-one mapping with the 3D heatmaps of hand joints compared to the occupancy grid representation.  
We believe that, in the occupancy grid representation, a lot of information is lost during the binary quantization of 3D depth point clould. Whereas, the projective TSDF representation better encapsulates this information which results in an improved hand pose estimation accuracy.  
This claim is confirmed by the results on the HANDS19 (Task 1) challenge, where HandVoxNet++ ranks first at the moment of our submission in August 2020. 
The runtime performance of the proposed registration components has significantly improved over \cite{HandVoxNet2020}. This improvement in GCN-MeshReg is mainly due to the fact that we do not need the time consuming hand shape surface voxelization step of DispVoxNet. 
{Our generative registration approach NRGA and its extended version NRGA++ are presented as an alternative to the learning-based registration. 
These generative registration approaches remove the dependency of the registration task on the availability of the shape annotations.
Although, such annotations are available in several recent datasets, they do not cover all possible hand shape variations.   
In this paper, the runtime of NRGA++ has significantly improved over NRGA specifically by using a pre-defined MANO hand model segmentation for the alignment.} 
All these improvements are reflected in the results on HANDS19 challenge, HO-3D and SynHand5M datasets. 
{For HO-3D dataset, HandVoxNet++ is able to produce reasonable pose and shape estimates under the challenging scenario of hands interacting with objects, although it has not been explicitly designed for this scenario.} 
We believe that our method brings us closer to in-the-wild AR and VR systems for hand shape estimation from depth sensors. 
HandVoxNet++ can regress over $30$ hand shapes per second in its fully neural configuration, and has potential for improvements by considering temporal shape context. 

{\noindent\textbf{Failure Cases.}} 
Our approach fails to estimate plausible hand shapes and poses in cases of severe occlusion of hand parts especially during hand-object interaction, see Fig.~\ref{fig:failure_dvn}. The difficulty in the estimation under such a scenario might also increase due to the large variation in object shapes. 

\section{Conclusion and Future Work}
We introduced a new HandVoxNet++ method for 3D hand shape and pose reconstruction from a single depth map, which %
establishes an effective inter-link between hand pose and shape estimations using 3D and graph convolutions. 
The experimental evaluation shows that the TSDF-based voxelized depth map representation establishes a more efficient one-to-one mapping with the 3D heatmaps of joint positions in comparison to the binary voxelized representation of the depth map. 
HandVoxNet++ produces more accurate hand shapes of real images compared to the previous methods, and our 3D data augmentation policy on voxelized grids further enhances the accuracy of 3D hand pose estimation. 
We achieve state-of-the-art results for 3D hand pose and shape estimation from depth images, which is confirmed on recent challenging benchmarks. 

All these results indicate that the one-to-one mapping between voxelized depth map, voxelized shape and 3D heatmaps of joints is essential for an accurate hand shape and pose recovery. 
As future work, the voxelized depth map can be combined with the color image to further enrich the voxelized input representation with an additional cue. 
Another promising direction is an extension of our work for reconstructing shapes of interacting hands. 
\ifCLASSOPTIONcompsoc
  \section*{Acknowledgments}
\else
  \section*{Acknowledgment}
\fi

This work was funded by the German Federal Ministry of Education and Research as part of the project VIDETE (Grant 01IW18002), DECODE (Grant 01IW21001), and the ERC Consolidator Grant 770784.

\ifCLASSOPTIONcaptionsoff
  \newpage
\fi

\bibliographystyle{IEEEtran}
\bibliography{egbib}

\vspace{-0.5cm}
\begin{IEEEbiography}[{\includegraphics[width=1in,height=1.25in,clip,keepaspectratio]{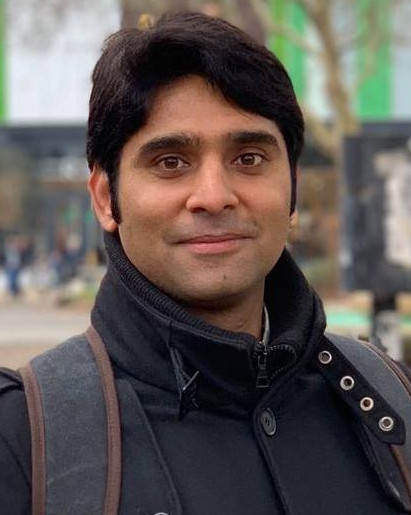}}]{Jameel Malik} is a postdoctoral researcher in the Augmented Vision group at the German Research Center for Artificial Intelligence (DFKI GmbH), Kaiserslautern. He received the PhD degree in computer science from Technische Universität Kaiserslautern in 2020 for his work on depth-based 3D hand pose and shape estimation, master’s degree in electrical engineering from the School of Electrical Engineering and Computer Science (SEECS),
National University of Sciences and Technology (NUST), Pakistan. His current research interests are in the areas of computer vision, deep learning and their applications.
\end{IEEEbiography}
\vspace{-1cm}
\begin{IEEEbiography}[{\includegraphics[width=1in,height=1.25in,clip,keepaspectratio]{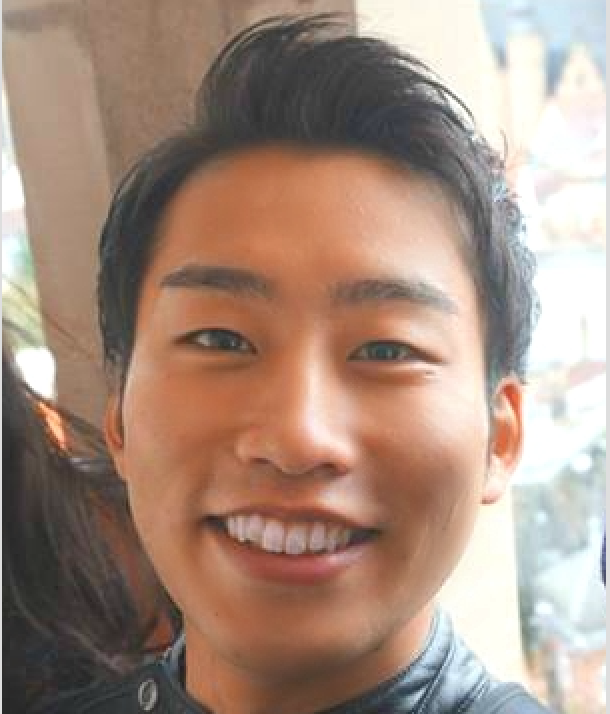}}]{Soshi Shimada}
is a Ph.D. candidate in the  Visual  Computing  and  Artificial  Intelligence Department  at the Max Planck Institute for Informatics in Saarbrucken, Germany.
Before, he received his M.Sc. in Computer Science from Technische Universität Kaiserslautern, and B.Eng. in Computer Science
and Engineering from Waseda University in Japan. His research interests are in computer vision, computer graphics, and machine learning, with a focus on 3D pose estimation. He currently works on monocular-based 3D human pose estimation.
\end{IEEEbiography}
\vspace{-1cm}
\begin{IEEEbiography}[{\includegraphics[width=1in,height=1.25in,clip,keepaspectratio]{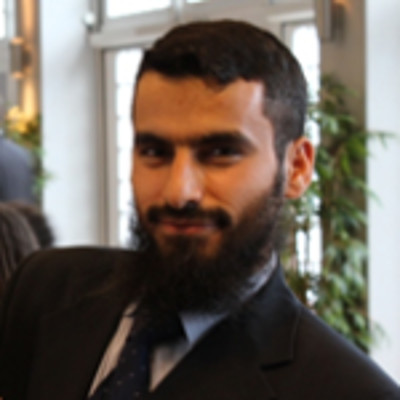}}]{Ahmed Elhayek} received the master's degree from Saarland University (Germany) in 2010. In 2015, he acquired his Ph.D. degree from both the Max-Planck-Institute and Saarland University. Thereafter, he worked as a PostDoc researcher in the Augmented Vision group at DFKI (German Research Centre for Artificial Intelligence). In 2018, he joined the faculty of Computer and Cyber Sciences in UPM (the University of Prince Mugrin) as an Assistant Professor.
\end{IEEEbiography}
\vspace{-1cm}
\begin{IEEEbiography}[{\includegraphics[width=1in,height=1.25in,clip,keepaspectratio]{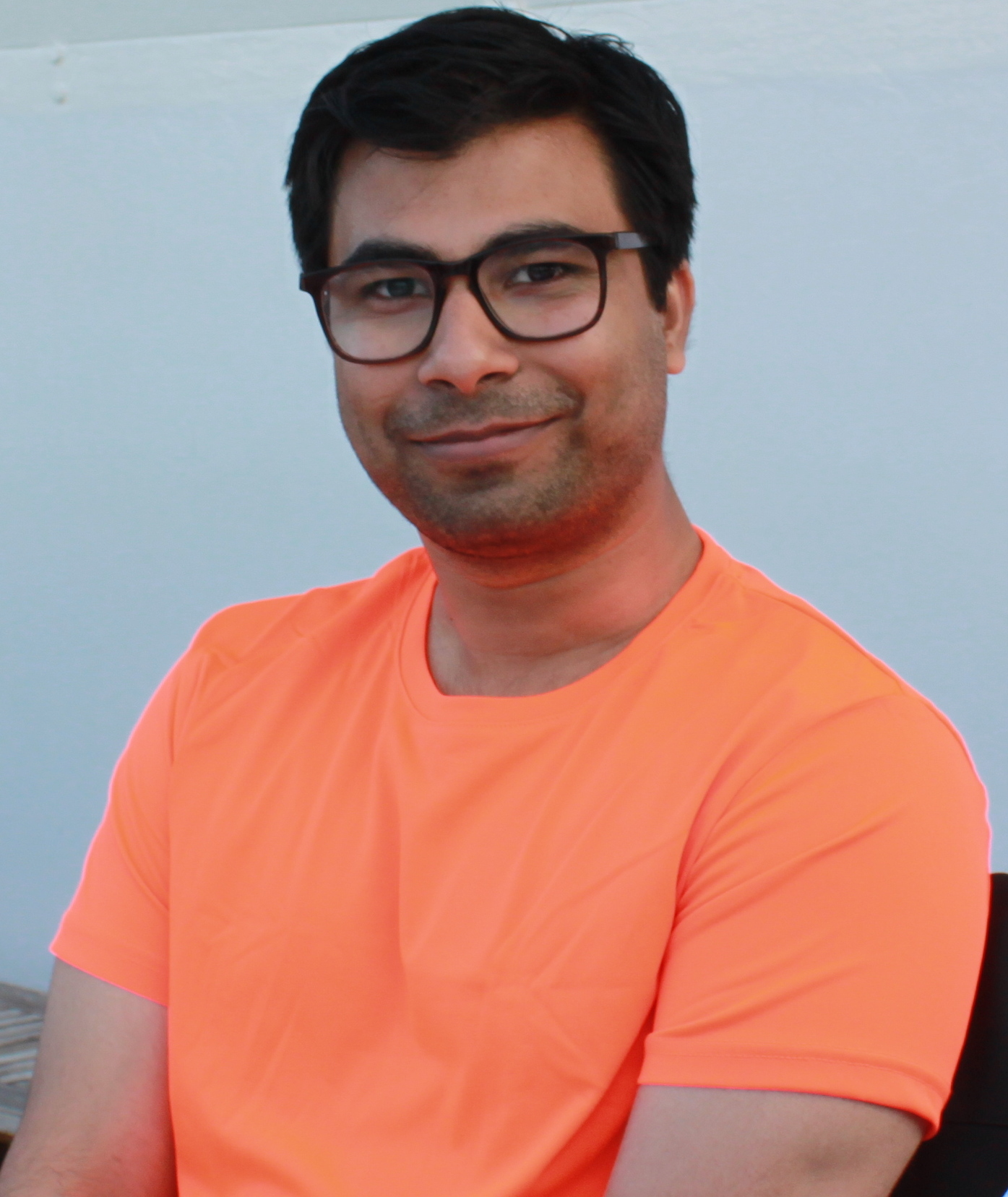}}]{Sk Aziz Ali}
is a researcher at the Augmented Vision group of German Research Center for Artificial Intelligence, Kaiserslautern.
He is doing the doctoral research at Technische Universität Kaiserslautern, in the areas of rigid and non-rigid motion
fields estimation, physics-based optimization methods and 3D reconstruction. He received the B.Tech degree from West
Bengal University of Technology and the M.Sc degree in computer science from Technische Universität Kaiserslautern in 2011 and 2017, respectively.
\end{IEEEbiography}
\vspace{-1cm}
\begin{IEEEbiography}[{\includegraphics[width=1in,height=1.25in,clip,keepaspectratio]{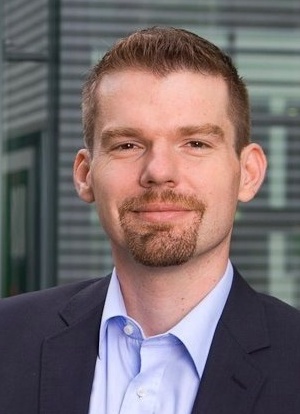}}]{Christian Theobalt}
is a Professor of Computer Science and the Director of the Visual Computing and Artificial Intelligence Department at the Max Planck Institute for Informatics, Saarbruecken, Germany. He is also a professor at Saarland University. His research lies at the Intersection of Computer Graphics, Computer Vision and Machine Learning. For instance, he works on virtual humans, 3D and 4D scene reconstruction, neural rendering and neural scene representations, marker-less motion and performance capture, machine learning for graphics and vision, and new sensors for 3D acquisition. Christian received several awards, for instance the Otto Hahn Medal of the Max-Planck Society (2007), the EUROGRAPHICS Young Researcher Award (2009), the German Pattern Recognition Award (2012), the EURIGRAPHICS Outstanding Technical Contributions Award (2020), an ERC Starting Grant (2013) and an ERC Consolidator Grant (2017). He is a co-founder of theCaptury (www.thecaptury.com). 
\end{IEEEbiography}
\vspace{-1cm}
\begin{IEEEbiography}[{\includegraphics[width=1in,height=1.25in,clip,keepaspectratio]{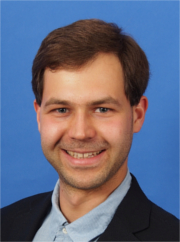}}]{Vladislav Golyanik} is leading the ``4D and Quantum Vision'' research group (4dqv.mpi-inf.mpg.de/) at the Visual Computing and Artificial Intelligence Department of the Max Planck Institute for Informatics (MPII), Saarbrücken, Germany. The primary research interests of his team include 3D reconstruction and analysis of deformable scenes, matching problems on point sets and graphs, neural rendering, quantum computer vision and event-based vision. He received a doctoral degree in informatics from the University of Kaiserslautern in 2019, advised by Didier Stricker. Prior to joining MPII as a post-doctoral researcher, Vladislav was a visiting fellow at NVIDIA (San Jos\'e, USA), and Institute of Robotics and Industrial Informatics (Barcelona, Spain). He is the recipient of the WACV’16 best paper award and the 2020's yearly dissertation award of the German Association for Pattern Recognition (DAGM). 
\end{IEEEbiography}
\vspace{-1cm}
\begin{IEEEbiography}[{\includegraphics[width=1in,height=1in,clip,keepaspectratio]{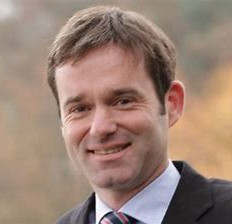}}]{Didier Stricker} is Professor in Computer Science at Technische Universität Kaiserslautern and Scientific Director at the German Research Center for Artificial Intelligence (DFKI GmbH) in Kaiserslautern, where he leads the research department 'Augmented Vision'. 
He received the Innovation Prize of the German Society of Computer Science in 2006.
He got several awards for best papers or demonstrations at different conferences.
He serves as reviewer for different European or National research organizations. He is a reviewer of different journals and conferences in the area of VR/AR and Computer Vision. His research interests are cognitive interfaces, user monitoring and on-body-sensor-networks, computer vision, video/image analytics, and human computer interaction. 
\end{IEEEbiography}

\end{document}